%% file: polaris1.tex
\documentclass[11pt]{article}

\usepackage[preprint]{acl}

\usepackage{times}
\usepackage{latexsym}

\usepackage[T1]{fontenc}

\usepackage[utf8]{inputenc}

\usepackage{microtype}

\usepackage{inconsolata}

\usepackage{graphicx}
\usepackage{amsmath}
\usepackage{amssymb}
\usepackage{booktabs}
\usepackage{multirow}
\usepackage{colortbl}
\usepackage{algorithm}
\usepackage[noend]{algpseudocode}

\usepackage{float}
\usepackage{placeins}
\usepackage{tcolorbox}
\usepackage{enumitem}
\usepackage{caption}
\usepackage{xcolor}
\usepackage{tikz}
\usetikzlibrary{positioning, arrows.meta}

\newcommand{\red}[1]{\textcolor{red}{#1}}
\newcommand{\sysname}{Polaris}

\newcommand{\daw}{AW}
\newcommand{\darctic}{Arctic}     
\newcommand{\dlter}{LTER}         
\newcommand{\decir}{ECIR}
\newcommand{\dwiki}{WikiTables}
\newcommand{\dwtr}{WTR}

\newcommand{\dsep}{\arrayrulecolor{black!25}\midrule\arrayrulecolor{black}}

\newtheorem{example}{Example}

\title{Polaris: Learning to Generate Table Descriptions from Retrieval Feedback}

\author{Ting Cai, Tuan Minh Phan, AnHai Doan \\
  University of Wisconsin-Madison \\
  \texttt{\{tingcai, minhrua, anhai\}@cs.wisc.edu} \\}

\begin{document}
\maketitle
\begin{abstract}
Many table-centric NLP tasks such as NL2SQL first retrieve relevant
tables from large collections using keyword search. Recent work uses
LLMs to generate natural-language table descriptions to improve
retrieval, but they are typically optimized for fluency rather than retrieval
effectiveness. We present Polaris, a system that trains an LLM to
generate table descriptions directly from retrieval feedback. Our key
insight is that existing table retrieval benchmarks already contain
the supervision needed for this task: given query--table relevance
judgments, we generate multiple candidate descriptions for each table,
rank them by their BM25 retrieval effectiveness, and use the resulting
preference pairs to fine-tune the LLM with Direct Preference
Optimization (DPO). Polaris further expands abbreviated table and column names
before generation to reduce vocabulary mismatch. Extensive experiments
show that Polaris outperforms the state-of-the-art AutoDDG solution,
often by a significant margin. More broadly, our results demonstrate
that retrieval benchmarks can be repurposed as supervision for
training LLMs to generate retrieval-oriented metadata.
\end{abstract}

\input{sections/intro}

\input{sections/problem_definition}

\input{sections/solution_architecture}

\input{sections/experiment}

\input{sections/related_work}

\input{sections/conclusion}
\input{sections/limitations}

\bibliography{custom}

\input{sections/appendix}

\end{document}

%% file: sections/intro.tex
\section{Introduction}\label{sec:intro}

Tabular data is ubiquitous in companies, sciences, government agencies,
and others \cite{zhang2023nameguess,edi}. As a result, the emerging direction of NLP
tasks for tabular data has grown significantly in the past ten years. Examples
of such tasks include NL2SQL, table QA, schema-based relation detection, 
and more \cite{liu2025survey,zhou2026tableqa,koutras2021valentine}.

Many of these tasks rely on a fundamental primitive, which is keyword
search over tables. This is because the set of tables under
consideration is often very large in practice. So given the user
intent, often expressed in natural language or keywords, we must first
perform keyword search (KWS) over these tables to find a
relatively small set of relevant tables, before we can perform
task-specific reasoning.  Consider for example NL2SQL. Given a natural
language query such as ``show quarterly sales for 2025'', many NL2SQL
solutions perform KWS to retrieve the relevant tables, then generate a
SQL query over these tables \cite{liu2025survey}.

\begin{figure}[t]
  {\footnotesize
\begin{verbatim}
ACTEMPS(EID, FN, LN, SAL)

This table stores the records of currently active
employees, where each row gives an employee's 
first name, last name, and salary. It supports
HR and payroll analyses such as staff compensation,
wage distribution, and headcount reporting.
\end{verbatim}
}
\caption{A tiny example of table description.}\label{ex1}
\end{figure}

While important, KWS over tables is difficult because the table and
column names are often cryptic and short (e.g., ``DTPh'',
``ActEmployees''), making it hard to search on. In response, many
recent works have proposed using LLMs to generate table descriptions in natural language,
then performing KWS over these descriptions to find relevant tables (see Figure \ref{ex1})
\cite{autoddg,pneuma2025,singh2025rag}. As far as we can tell, the most advanced work
in this direction is AutoDDG, which is shown empirically
to outperform six other solutions \cite{autoddg}. While promising,
the AutoDDG work has several major limitations, as we discuss below. In this
paper we describe Polaris, which addresses these limitations
and significantly advances the state of the art.

First, AutoDDG employs a relatively complex pipeline with multiple LLM
calls to analyze a table's metadata (e.g., table name, column names)
and tuples before generating its description. We show that a much
simpler pipeline, requiring only a single LLM call, achieves
comparable or even better performance, and we explain why.

Second, AutoDDG uses the table names and column names ``as is''.  In
practice, these names are often cryptic or abbreviated, e.g.,
\texttt{DTPh} for ``daytime phone'' \cite{columbo}.  Such names give the LLM little
to work with, producing generic descriptions that miss important
vocabulary. We show that by expanding these names into full English
phrases, using the solution in \cite{columbo}, we can provide richer
input to the LLM and can substantially improve the quality of the
generated descriptions.

Third, AutoDDG uses the LLM ``as is'', without fine tuning.  We show
how to fine tune the LLM to produce better table descriptions for
keyword search. Our approach is inspired by reinforcement learning
with human feedback (RLHF) \cite{ziegler2019fine}, in which a provider such as
ChatGPT asks the user to rank alternative answers to the same question, uses
the rankings to learn a reward function, then uses the reward function
in a reinforcement learning process to fine tune the LLM to produce
better answers.

We adapt this paradigm in two ways. First, we do not have
access to many human users, and even if we do, they typically cannot tell
which table description is better for KWS. So we use labeled data, in
which given a set of tables and a set of keyword queries, we know
which tables are relevant to which queries. Several such labeled
datasets already exist, and are widely used to train and evaluate
KWS solutions (see the experiment section). We show how to use these
labeled datasets to rank table descriptions. Then instead of RLHF, we
use the rankings to fine tune the LLM via direct preference
optimization (DPO), a recent popular technique \cite{dpo}.

The fourth limitation is that AutoDDG uses only the generated table
descriptions in KWS over tables. We show that, while useful, these
descriptions are not perfect, so using them {\em together\/} with
other table metadata (e.g., table/column name expansions,
contexts, etc.)  can significantly improve KWS accuracy.

Finally, the AutoDDG work uses only two labeled datasets for
evaluation. We have cleaned 3 major labeled datasets that have been
used in research on KWS over tables, and created 3 new labeled
datasets. This produces a benchmark of 6 datasets consisting of
(table, query, relevance score) triplets. As far as we can tell, this
is the largest benchmark to date for evaluating KWS over tables. In
summary, we make the following contributions:
\begin{itemize}
\item We develop Polaris, a solution that significantly improves upon AutoDDG. Polaris
  uses a simpler pipeline to generate table descriptions, expands cryptic table and column
  names into full English phrases, uses labeled data to fine tune LLMs to produce better
  descriptions, and exploits these descriptions as well as other table metadata for KWS.

\item We develop a benchmark of six labeled datasets, the largest benchmark to date
  for evaluating KWS solutions over tables.

\item Extensive experiments show that Polaris outperforms AutoDDG
  on the above benchmark, often by a large margin (e.g., up to 44.6\% increase
  of NDCG@100). Further, Polaris generates much shorter table descriptions than AutoDDG,
  reducing the average description length by 47-59\%.
\end{itemize}
More broadly, our results suggest that retrieval benchmarks can be
reused as supervision for training LLMs to generate
retrieval-oriented metadata.
The code is available at github.com/anhaidgroup/polaris, and the
datasets at hf.co/collections/anhaidgroup/polaris-v1.

%% file: sections/problem_definition.tex
\section{Preliminaries}\label{sec:problem}
In this section we describe AutoDDG then build on that to describe
the problem setting of Polaris.

\paragraph{AutoDDG:} This solution uses
a complex multi-step pipeline to generate table descriptions
\cite{autoddg}. For a particular table $T$, the pipeline takes as input
the table name and column names, additional metadata if any, sample tuple
values (if available). It generates table
statistics, table topics, and column summaries, then uses these to
generate a table description $d(T)$ geared toward
KWS. This pipeline makes at least four LLM calls, often many more.

Given a keyword query $Q$, AutoDDG performs
KWS to return a ranked list of tables. Here each table $T$ is ranked
based on a similarity score between its description $d(T)$ and query
$Q$. This similarity score is the BM25 TF/IDF score widely used in
today's KWS engines \cite{robertson2009bm25}.

The paper \cite{autoddg} shows that AutoDDG outperforms six solutions,
including two that are pre-trained data-to-text models
~\cite{mvp,reastap}, and
Pneuma~\cite{pneuma2025}, which uses LLMs to summarize the schema and
sampled rows for RAG-style table discovery.

\begin{figure*}[t]
\centering
\includegraphics[width=\textwidth]{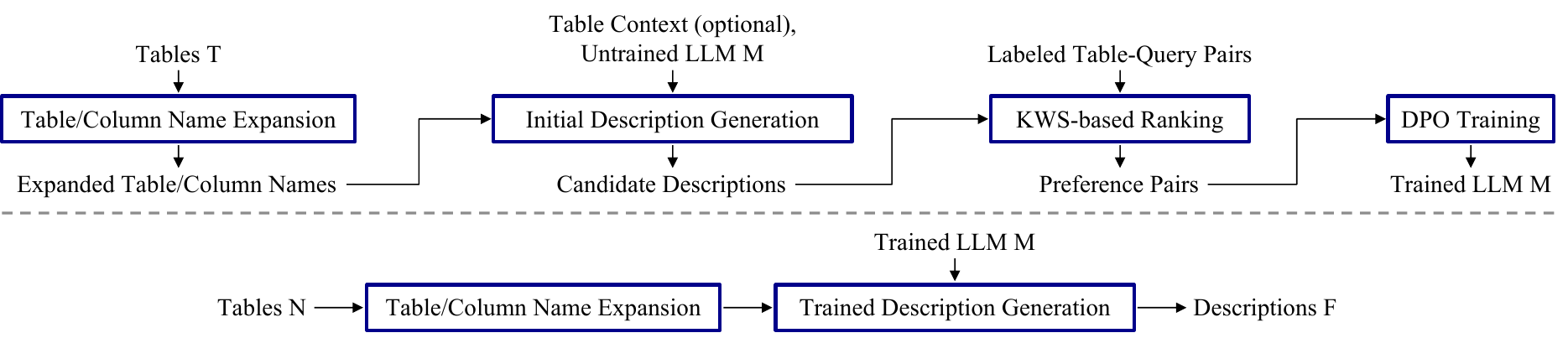}
\caption{The overall architecture of Polaris.}
\label{arch}
\end{figure*}

\paragraph{Table Description Generation in Polaris:} Similar to AutoDDG,
to generate a description $d(T)$ for a table $T$, we
also use its metadata: table name, column names, and
additional context if any (e.g., table caption, Web page title).

Unlike AutoDDG, however, we do not use the tuples of $T$ in
description generation. Such tuples are often not available, for privacy and legal reasons. Furthermore, we
show later that even without using the tuples Polaris already
outperforms AutoDDG with tuples (see Section~\ref{sec:experiments}).

As discussed in the introduction, another distinguishing aspect of
Polaris is that we use labeled datasets, each of which consists of
tables, queries, and relevance scores, to fine tune the LLM to produce
better descriptions for KWS (see Section~\ref{sec:method}).

\paragraph{Keyword Search in Polaris:} We adopt the same KWS setting as AutoDDG.
Specifically, given a keyword query $Q$, we
return a ranked list of tables, where each table $T$ is ranked
based on the popular BM25 TF/IDF score between its description
$d(T)$ and query $Q$. Considering other similarity scores (e.g.,
embedding vectors, hybrid \cite{karpukhin2020dpr,bruch2023fusion,formal2021splade}) is left for future work.

Since the generated table descriptions are not perfect (see
Section~\ref{sec:experiments}), unlike AutoDDG, we will also consider
a KWS setting where the BM25 score uses both the descriptions and
other table metadata. We show that this new score can significantly
improve KWS accuracy.

%% file: sections/solution_architecture.tex
\section{The Polaris Solution}\label{sec:method}

We now describe how Polaris generates table descriptions in two
stages: {\em fine tuning an LLM\/} and {\em using the LLM to generate
  descriptions}.

In the first stage (see the top part of Figure \ref{arch}), given a
set of tables (used for training, as we discuss later), we expand the
table/column names into full English phrases (e.g., ``eSal'' into
``employee salary''). An untrained LLM then uses these expansions and table contexts (if any) to
generate for each table multiple candidate 
descriptions. Next, we use labeled datasets to rank these
descriptions (regarding their quality for KWS over tables).  Finally,
we use these rankings to train, i.e., fine tune the LLM, using
direct preference optimization (DPO).

In the second stage (see the bottom part of Figure \ref{arch}), given
a corpus of tables $N$ (over which we want to enable KWS), we expand
the table/column names. The trained LLM then uses these expansions and
table contexts (if any) to generate for each table a description. We
now describe the main components of Polaris in detail.

\subsection{Table/Column Name Expansion}\label{sec:expansion}
We assume Polaris has access to a set of labeled datasets $\mathcal{D}
= \{D_1,\ldots, D_n\}$.  Each dataset consists of (table, query,
relevance score) triplets. Higher score indicates more relevance, and
a score of 0 indicates no relevance. Several such labeled datasets
already exist and have been widely used in research on KWS over
tables.

Let $\mathcal{T}$ be the set of all tables in $\mathcal{D}$. Later we
will generate alternative descriptions for these tables, then rank
them. To generate good descriptions, here we will expand the
table/column names in $\mathcal{T}$ into English phrases (e.g.,
``DTPh'' to ``daytime phone'', ``russ\_ind'' to ``russell
index''). This is because the original names are often short and
cryptic. So when fed these names, the LLM often produces vague descriptions, hurting KWS.

To expand the table/column names, we use Columbo \cite{columbo}, a
state-of-the-art solution that uses an LLM on the original
names. Appendix~\ref{sec:app-expansion} describes how we adapt Columbo
to the Polaris context.

\subsection{Initial Table Description Generation}\label{sec:generation}

Next, we generate descriptions for the tables in the labeled datasets.
For each table $T$, we send a prompt to an LLM $M$ (with no fine
tuning). The prompt provides $M$ with the table/column name
expansions, the context (if any), and an one-shot example adapted
from~\cite{autoddg}.

Inspired by the descriptions in AutoDDG, we ask the LLM to produce
similar descriptions, which are JSON objects with six fields: data set
overview, key themes/topics, applications/use cases,
concepts/synonyms, keywords/themes, and additional context.  See
Appendix~\ref{sec:app-initial-generation} for the prompt and a sample
description.

Recall that each labeled dataset consists of (table, query, score)
triplets. If a table has a relevance score greater than 0, we call it
a \underline{gold table}, because it is a ``gold'' answer to at least one
query. For each gold table we generate $k \geq 2$ table
descriptions, by sampling with non-zero temperature and top-p
sampling, so that the LLM produces diverse outputs from the same
prompt. For each remaining (non-gold) table, we generate a
single description. The reason why we treat these two kinds
of tables differently will be clear in the next subsection.

\begin{figure}[t]
\centering
\includegraphics[width=\columnwidth]{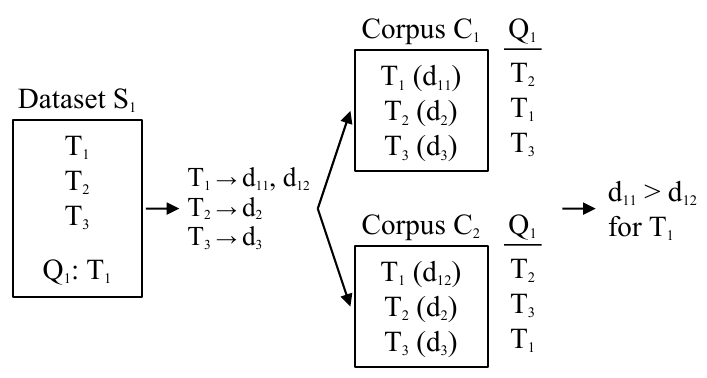}
\caption{Using a labeled dataset to generate rankings.}
\label{rank-ex}
\end{figure}

\subsection{Ranking the Table Descriptions}\label{sec:ranking}

We now rank the table descriptions. Human users typically cannot
tell which description is better for a table
KWS-wise. So our key idea is to use the labeled datasets to rank. The
following example illustrates this idea.
\begin{example}
Consider the labeled dataset $S_1$ in Figure \ref{rank-ex}, which has
3 tables $T_1,T_2,T_3$ and 1 query $Q_1$.  Suppose $T_1$ is the only
``gold'' answer to $Q_1$ (i.e., having a relevance score greater than
0).  Suppose that we have used an LLM to generate 2 descriptions
$d_{11}, d_{12}$ for the gold table $T_1$ and 1 description each for
the non-gold tables $T_2$ and $T_3$ (see Figure \ref{rank-ex}).

Then our goal is to rank descriptions $d_{11}$ and $d_{12}$, i.e.,
deciding which one is better KWS-wise for the gold table $T_1$. To do
so, we construct two corpora $C_1$ and $C_2$, as shown in Figure
\ref{rank-ex}.  These corpora are identical, except that table
$T_1$ uses different descriptions: $d_{11}$ in $C_1$, and $d_{12}$ in
$C_2$. All other tables are non-gold, each of them uses the single
description that we have generated.

Suppose performing query $Q_1$ on corpus $C_1$ and corpus $C_2$
produces the ranking $R_1 = \langle T_2, T_1, T_3\rangle$ and $R_2 =
\langle T_2, T_3, T_1\rangle$, respectively (see Figure \ref{rank-ex}).
Then $T_1$ is ranked higher in $R_1$ compared to $R_2$, suggesting
that in this case description $d_{11}$ is better than description
$d_{12}$ for table $T_1$.
\end{example}
We realize the above idea as follows. First, for each gold table we combine
all of its corpora into one corpus. For example, for gold table $T_1$ in Figure \ref{rank-ex},
we combine the two corpora $C_1$ and $C_2$ into a single corpus $C_{12} = \{T_1(d_{11}),
T_1(d_{12}), T_2(d_2), T_3(d_3)\}$. This corpus has 2 copies for table $T_1$, each with
a different description, and 1 copy for each remaining table, each with a randomly
selected description.

Second, we score the two descriptions of the gold table and rank them
based on their scores. Continuing with the above example, the score of
description $d_{11}$ (for table $T_1$) is the BM25 score between query
$Q_1$ (for which $T_1$ is a gold answer) and $d_{11}$. If $T_1$ is a
gold answer for {\em multiple queries\/}, then the score for $d_{11}$ is the
{\em average\/} of its scores for these queries.  We compute the
score of description $d_{12}$ in a similar fashion.

Suppose the score of $d_{11}$ is greater than that of
$d_{12}$. Intuitively this means that for the queries for which $T_1$
is a correct answer, using description $d_{11}$ will place $T_1$
higher in the ranked list of tables, compared to using description
$d_{12}$. As a result, we rank $d_{11}$ higher, denoted as $d_{11} >
d_{12}$.

Finally, we generate 3 descriptions for each gold table, not 2.  For
example, for gold table $T_1$, we generate $d_{11}, d_{12}, d_{13}$.
Now suppose we can rank $d_{13} > d_{11} > d_{12}$, as previously
discussed.  Then we keep only the ranking that we judge to be the most
reliable, namely $d_{13} > d_{12}$, to be used in fine tuning the
LLM. (Thus, we do not use the other two rankings $d_{13} > d_{11},
d_{11} > d_{12}$.)

Algorithm \ref{alg:polaris-training} provides the pseudo code of our
complete solution. As discussed earlier, this solution uses one
``unified'' corpus for each gold table, not multiple ``individual''
corpora. Using multiple corpora incurs a significant cost in creating
the corpora and indexing them for querying. So we approximate this
process with just one corpus. Since this ``unified'' corpus is
identical to each ``individual'' corpus, except that it has 2 extra
copies for the target gold table, we found the approximation to
produce negligible loss in BM25 scores.

Further, our solution assigns a {\em random\/} description to
each non-target table in the corpus (e.g., assigning $d_2$ to $T_2$
and $d_3$ to $T_3$ in Figure \ref{rank-ex}). Ideally, we should run
this solution multiple times. Each time we assign the {\em best\/}
description (i.e., the one with top rank) found so far to each non-target table
in the corpus.

For example, suppose we have determined $d_{11} > d_{12}$ for $T_1$ (see Figure \ref{rank-ex}). Now suppose that $T_2$
is also a gold table. Then when creating a corpus for $T_2$, we should
assign description $d_{11}$ to $T_1$. We should repeat this process
until a convergence condition. Doing this however is very time consuming.
As a result, currently we run the above solution only once for each gold
table (in which all remaining tables in the same corpus are assigned
random descriptions). The experiments show that this is already effective
in generating good descriptions for KWS. Exploring running the above
solution multiple times is ongoing work.

\begin{algorithm}[t!]
\footnotesize
\renewcommand{\algorithmicindent}{1em}
\caption{The training procedure of \sysname{}.}
\label{alg:polaris-training}
\begin{algorithmic}[1]
\Require
\Statex \hspace{1.5em}$\mathcal{D}$ --- labeled datasets, each $D = (\mathcal{T}, \mathcal{Q}, R)$:
\Statex \hspace{3em}$\mathcal{T}$ --- tables
\Statex \hspace{3em}$\mathcal{Q}$ --- keyword queries
\Statex \hspace{3em}$R:\mathcal{Q}{\times}\mathcal{T}\to\{0,1,\dots\}$ --- relevance judgments
\Statex \hspace{4.5em}($0$ = not relevant)
\Statex \hspace{1.5em}$\mathcal{M}$ --- untrained LLM
\Statex \hspace{1.5em}$K$ --- number of candidate descriptions per table
\Ensure
\Statex \hspace{1.5em}$\mathcal{M}^{*}$ --- trained LLM
\State $P \gets \emptyset$ \Comment{preference pairs}
\For{each dataset $D \in \mathcal{D}$}
    \State Expand table and column names in $D$
    \State $G \gets$ tables with a relevant query ($\exists q\colon R(q,T)>0$)
    \State $C_0 \gets \emptyset$ \Comment{default corpus}
    \For{each table $T \in \mathcal{T}$}
        \If{$T \in G$}
            \State Generate $K$ candidate descriptions $C_T$ with $\mathcal{M}$
            \State Add a random description from $C_T$ to $C_0$
        \Else
            \State Generate one description with $\mathcal{M}$
            \State Add it to $C_0$
        \EndIf
    \EndFor
    \For{each table $T \in G$}
        \State $\mathcal{C} \gets C_0 \cup C_T$ \Comment{table $T$'s corpus}
        \For{each candidate description $c \in C_T$}
            \State $s(c) \gets$ mean $\mathrm{BM25}_{\mathcal{C}}(q,c)$ over relevant queries
        \EndFor
        \State $c_{\mathrm{best}} \gets \arg\max_{c \in C_T} s(c)$ \Comment{best description}
        \State $c_{\mathrm{worst}} \gets \arg\min_{c \in C_T} s(c)$ \Comment{worst description}
        \State Add $(\mathrm{prompt}(T),\ c_{\mathrm{best}},\ c_{\mathrm{worst}})$ to $P$
    \EndFor
\EndFor
\State $\mathcal{M}^{*} \gets \mathcal{M}$ trained with DPO on $P$
\State \Return $\mathcal{M}^{*}$
\end{algorithmic}
\end{algorithm}

\subsection{DPO Training with the Rankings}\label{sec:training}

We now use the rankings of the descriptions to fine tune the base LLM via DPO.
Recall that in the previous steps, for each gold table $T$, we have generated
and ranked 3 descriptions $d_{\pi_1} > d_{\pi_2} > d_{\pi_3}$. This yields
the following preference pair: 
\[
\bigl(\,\text{prompt}(T),\;\; d_{\text{win}} = d_{\pi_1},\;\; d_{\text{lose}} = d_{\pi_3}\,\bigr),
\]
where $\text{prompt}(T)$ is the prompt used for generating the description of table $T$.

Let $P$ be the set of all preference pairs, one per gold table (from
all labeled datasets).  We then fine tune the base LLM $M$ using
direct preference optimization (DPO)~\cite{dpo}, which trains the
LLM to increase the likelihood of~$d_{\text{win}}$ relative
to~$d_{\text{lose}}$ without a separate reward model.  The DPO loss
is $\mathcal{L}_{\text{DPO}} = -\mathbb{E}\![\log\sigma\!(
\beta\log\frac{\pi_\theta(d_{\text{win}} \mid x)}{\pi_{\text{ref}}(d_{\text{win}} \mid x)}
- \beta\log\frac{\pi_\theta(d_{\text{lose}} \mid x)}{\pi_{\text{ref}}(d_{\text{lose}} \mid x)})]$,
where $\pi_\theta$ is the model being trained,
$\pi_{\text{ref}}$ is the frozen base model,
$x$ is the prompt,
and $\beta$ controls regularization strength.

We use Llama 3.1 8B Instruct as the base model, with 4-bit
quantization and LoRA~\cite{lora} for parameter-efficient
fine-tuning. The adapter targets all attention and feed-forward
projections (\texttt{q\_proj}, \texttt{k\_proj}, \texttt{v\_proj},
\texttt{o\_proj}, \texttt{gate\_proj}, \texttt{up\_proj}, and
\texttt{down\_proj}), with rank $r=64$ and $\alpha=64$.  This gives
167M trainable parameters out of 8.2B total (2.05\%).  We train for one
epoch with $\beta=0.1$, learning rate $5\times10^{-6}$, a linear
schedule with 10\% warmup, effective batch size~8, and the AdamW 8-bit
optimizer. See Appendix~\ref{sec:app-dpo} for details.

\subsection{Description Generation for New Tables}\label{sec:inference}

Once fine tuned, the LLM $M$ is ready to use. Given a new corpus of
tables on which we want to perform KWS, for each table in the corpus,
we first expand the table/column names, then feed
these together with other metadata into the LLM $M$ to generate a
table description, using the same prompt described in Section~\ref{sec:generation}.

%% file: sections/experiment.tex
\section{Experiments}\label{sec:experiments}

\begin{table}[t]
\centering
\footnotesize
\begin{tabular}{llrrr}
\toprule
Dataset & Domain & \#Tables & \#Queries & \#Gold \\
\midrule
\daw{}         & Enterprise  & 96    & 15 & 41    \\
\darctic{}     & Science     & 251   & 20 & 121   \\
\dlter{}       & Science     & 2,015 & 15 & 374   \\
\decir{}       & Government  & 2,100 & 12 & 1,411 \\
\dwtr{}        & Web         & 4,634 & 60 & 2,190 \\
\dwiki{}       & Web         & 3,361 & 57 & 845   \\
\bottomrule
\end{tabular}
\caption{Datasets for our experiments. \#Gold is
the number of gold tables.}
\label{datasets}
\end{table}

We now evaluate Polaris, using the six labeled datasets described in
Table \ref{datasets}. We created the first three datasets, while the
last three have been used in prior research on KWS over tables. Appendix~\ref{sec:app-datasets} describes
these datasets, as well as the LLM, DPO, and KWS settings
of our experiments.

We perform each experiment in a {\em leave-one-out\/}
fashion.  For example, for the AW dataset, Polaris fine tunes the
LLM {\em using the remaining five labeled datasets}, 
uses it to generate descriptions for the tables in AW, 
performs KWS for the queries in AW (using the generated
descriptions), and reports the aggregated accuracies over these queries
(using their gold tables). We consider two accuracy measures widely
used in KWS research: NDCG@K (ranking quality with graded relevance)
and Recall@K \cite{jarvelin2002cumulated,manning2008ir}.

\paragraph{Overall Performance:} Tables \ref{ndcg}-\ref{recall}
compare Polaris with AutoDDG in terms of NDCG@K and Recall@K,
respectively. They show that Polaris outperforms AutoDDG on the first
five datasets for nearly all $K$-s and for both measures, often by a
significant margin (e.g., NDCG@100 on LTER: $0.356 \to 0.515$,
$+44.6\%$ relative, and NDCG@5 on AW: $0.555 \to 0.731$, $+31.7\%$
relative). On the last dataset, WikiTables, both methods perform
similarly (e.g., AutoDDG leads at R@100 by $0.02$ while Polaris leads
at NDCG@10 by $0.01$).

Interestingly, we found that Polaris generates much shorter
table descriptions than AutoDDG, reducing the average
description length by 47-59\% (see Appendix~\ref{sec:overall}).

\begin{table}[t]
\centering
\footnotesize
\resizebox{\columnwidth}{!}{%
\begin{tabular}{llccccc}
\toprule
Dataset & Method & @5 & @10 & @20 & @50 & @100 \\
\midrule
\daw{} & AutoDDG & .5551 & .6192 & .6692 & .6969 & .7030 \\
       & \sysname{} & \textcolor{blue}{.7312} & \textcolor{blue}{.7582} & \textcolor{blue}{.7869} & \textcolor{blue}{.7869} & \textcolor{blue}{.7869} \\
\dsep
\darctic{} & AutoDDG & .6274 & .5942 & .5942 & .6360 & .6812 \\
           & \sysname{} & \textcolor{blue}{.7133} & \textcolor{blue}{.6743} & \textcolor{blue}{.6939} & \textcolor{blue}{.7416} & \textcolor{blue}{.7678} \\
\dsep
\dlter{} & AutoDDG & .4372 & .4194 & .4011 & .3656 & .3559 \\
         & \sysname{} & \textcolor{blue}{.7218} & \textcolor{blue}{.6546} & \textcolor{blue}{.6013} & \textcolor{blue}{.5239} & \textcolor{blue}{.5145} \\
\dsep
\decir{} & AutoDDG & .4215 & .4429 & .4512 & .4408 & .4180 \\
         & \sysname{} & \textcolor{blue}{.4946} & \textcolor{blue}{.5050} & \textcolor{blue}{.4981} & \textcolor{blue}{.4873} & \textcolor{blue}{.4672} \\
\dsep
\dwtr{} & AutoDDG & .3748 & .3548 & .3767 & .4159 & .4884 \\
        & \sysname{} & \textcolor{blue}{.3840} & \textcolor{blue}{.3840} & \textcolor{blue}{.3894} & \textcolor{blue}{.4449} & \textcolor{blue}{.5309} \\
\dsep
\dwiki{} & AutoDDG & \textcolor{blue}{.4139} & .4085 & .4398 & \textcolor{blue}{.5176} & \textcolor{blue}{.5811} \\
         & \sysname{} & .4111 & \textcolor{blue}{.4215} & \textcolor{blue}{.4407} & .5171 & .5727 \\
\bottomrule
\end{tabular}%
}
\caption{NDCG@K of AutoDDG vs. Polaris}
\label{ndcg}
\end{table}

\begin{table}[t]
\centering
\footnotesize
\resizebox{\columnwidth}{!}{%
\begin{tabular}{llccccc}
\toprule
Dataset & Method & @5 & @10 & @20 & @50 & @100 \\
\midrule
\daw{} & AutoDDG & .49 & .64 & .76 & \textcolor{blue}{.83} & \textcolor{blue}{.85} \\
       & \sysname{} & \textcolor{blue}{.66} & \textcolor{blue}{.77} & \textcolor{blue}{.83} & \textcolor{blue}{.83} & .83 \\
\dsep
\darctic{} & AutoDDG & .24 & .34 & .48 & .68 & .77 \\
           & \sysname{} & \textcolor{blue}{.29} & \textcolor{blue}{.41} & \textcolor{blue}{.58} & \textcolor{blue}{.77} & \textcolor{blue}{.82} \\
\dsep
\dlter{} & AutoDDG & .08 & .14 & .22 & .32 & .35 \\
         & \sysname{} & \textcolor{blue}{.18} & \textcolor{blue}{.25} & \textcolor{blue}{.32} & \textcolor{blue}{.39} & \textcolor{blue}{.43} \\
\dsep
\decir{} & AutoDDG & .01 & .02 & .04 & .08 & .12 \\
         & \sysname{} & .01 & \textcolor{blue}{.03} & .04 & .08 & \textcolor{blue}{.13} \\
\dsep
\dwtr{} & AutoDDG & .08 & .14 & .25 & .43 & .59 \\
        & \sysname{} & .08 & \textcolor{blue}{.15} & \textcolor{blue}{.26} & \textcolor{blue}{.48} & \textcolor{blue}{.66} \\
\dsep
\dwiki{} & AutoDDG & .21 & .29 & .41 & .66 & \textcolor{blue}{.83} \\
         & \sysname{} & .21 & \textcolor{blue}{.32} & \textcolor{blue}{.42} & \textcolor{blue}{.67} & .81 \\
\bottomrule
\end{tabular}%
}
\caption{Recall@K of AutoDDG vs. Polaris.}
\label{recall}
\end{table}

\paragraph{Ablation Studies:} We now evaluate the major components of Polaris.

\noindent{\em Prompt Quality:} First, we disable DPO training and
table/column name expansion, by modifying the prompt (see
Appendix~\ref{sec:app-initial-generation}) to use only the original
table/column names, then use this prompt on the base LLM (without fine
tuning) to generate table descriptions. The results show that the
``disabled'' Polaris still substantially outperforms AutoDDG on 4
datasets (AW, LTER, Arctic, ECIR), and performs similarly on 2
datasets (WTR, WikiTables). See
Tables~\ref{tab:prompt-ndcg}-\ref{tab:prompt-recall} in
Appendix~\ref{sec:app-ablation}. This suggests our simpler prompt can
already perform better or competitive with AutoDDG.

This is likely because of the multi-step ``lossy'' nature of AutoDDG's
prompt.  Specifically, AutoDDG (1) uses the LLM to generate column
summaries, table topics, etc., then (2) feeds these into the LLM again
to generate a table description. Step 1 is ``lossy'' due to the
imperfection of the LLM and the length limitations of the
summaries/topics. So the description generated from the output
of Step 1 is not as accurate as it can be. 

\noindent{\em Effect of DPO Training:}
Tables~\ref{tab:ablation-ndcg}-\ref{tab:ablation-recall} in
Appendix~\ref{sec:app-ablation} compare the full Polaris with
``P-DPO'', the Polaris version without DPO training (i.e., base LLM
with name expansion). The tables show that DPO training improves
\daw{} ($+9.4\%$ at @5), \dlter{} ($+10.4\%$ at @5), and \dwtr{}
($+5.0\%$ at @5).  The exception is \decir{}, where accuracy decreases
(e.g., $.495$ vs.\ $.570$ at @5). This is probably due to the very
high density of relevant tables per query: almost any reasonable
description retrieves a large fraction of relevant tables, so the
preference signal from DPO adds little in this low-discrimination
regime.

\noindent{\em Effect of Table/Column Name Expansion:}
Tables~\ref{tab:ablation-ndcg}-\ref{tab:ablation-recall} in
Appendix~\ref{sec:app-ablation} compare the full Polaris with
``P-exp'', the Polaris version without name expansion (i.e., using
DPO-trained LLM, with only original table/column names). The tables
show that expansion improves \dlter{} ($+21.2\%$ at @5), \daw{}
($+5.4\%$ at @5), \darctic{} ($+6.6\%$ at @5), and \dwtr{} ($+8.0\%$
at @5).  The benefit for \decir{} and \dwiki{} (which rely on external
context rather than table names) exists but is small.

\paragraph{Sensitivity Analysis:} Next we examine how Polaris's accuracy
changes regarding four major design choices: the LLM, the
DPO regularization parameter $\beta$, the number of candidate
descriptions per table, and the number of training epochs. For
space reasons we only briefly summarize the findings here (see
Appendices~\ref{sec:app-sens-llm}-\ref{sec:app-sens-epochs} for detail).

Regarding LLMs, we replace Llama 3.1 8B Instruct with
Qwen2.5-7B-Instruct for both Polaris and AutoDDG, keeping all other
settings identical. Table~\ref{tab:sens-llm-polaris} shows that
Polaris with Qwen outperforms Polaris with Llama on
\daw{}, \darctic{}, and \dwiki{}, but underperforms on \dlter{}, while
\decir{} and \dwtr{} show mixed results across
cut-offs. Table~\ref{tab:sens-llm-vs} shows that Polaris with Qwen
outperforms AutoDDG with Qwen on all datasets, confirming that Polaris remains superior to AutoDDG
under a different LLM.

Next, we increase the DPO regularization parameter $\beta$ from 0.1
(default) to 0.5, making the DPO objective more conservative (closer
to the reference model). Tables~\ref{tab:sens-beta}-\ref{tab:sens-beta-recall} show that Polaris with $\beta
= 0.5$ improves \daw{} (the smallest dataset) but hurts \dlter{} and
has negligible effect elsewhere.  The default $\beta = 0.1$ provides a
better balance across datasets.

Next, we increase the number of candidate descriptions per gold table
from $K=3$ (default) to $K=5$. Even with $K=5$, we still select only the best and worst performing
descriptions and form one training pair per table.
Tables~\ref{tab:sens-ndesc}-\ref{tab:sens-ndesc-recall} show that the new Polaris
substantially improves \daw{} ($+9.7\%$ at @5)
and \dwiki{} ($+5.3\%$ at @5)
but hurts \dlter{} ($-11.8\%$ at @5) and \darctic{} ($-4.1\%$ at @5).
More candidates provide a wider range of preference pairs
but may also introduce noise when the best and worst candidates
are harder to distinguish.

Finally, we increase training from 1 epoch (default) to 3 epochs.
Tables~\ref{tab:sens-epochs}-\ref{tab:sens-epochs-recall} show that the new Polaris substantially improves
\darctic{} ($+13.3\%$ at @5) but hurts \decir{}, \dwtr{}, and
\dwiki{}.  \daw{} shows mixed results (worse at early cut-offs, better
at deep ones).  Additional training helps datasets with many gold
tables in training (\darctic{} contributes many preference pairs) but
risks overfitting on smaller training sets, supporting our default of
1 epoch.

\begin{figure}[t]
\centering
\includegraphics[width=\columnwidth]{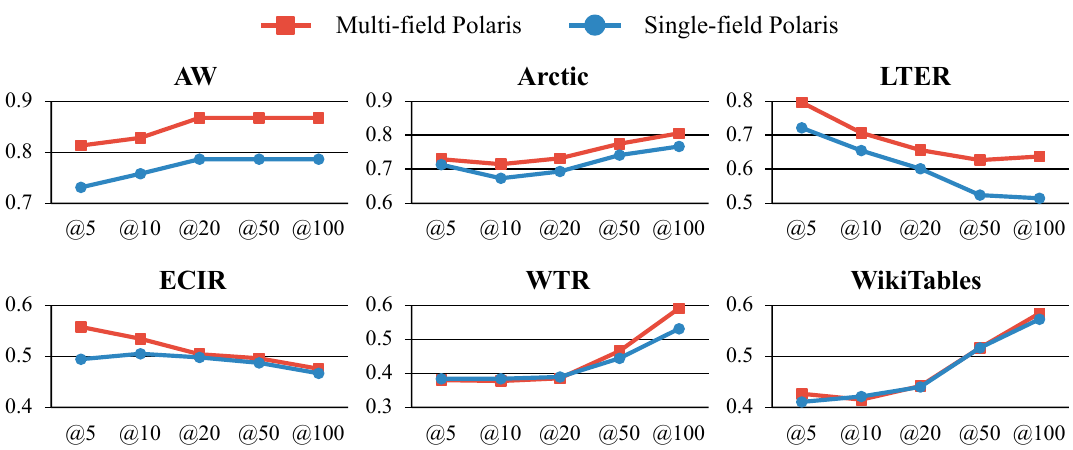}
\caption{NDCG@K of multi-field Polaris vs. single-field Polaris.}
\label{multifields}
\end{figure}

\paragraph{Multi-Field KWS with Polaris:} So far Polaris uses
only the generated table descriptions for KWS. Since these
descriptions are not perfect, we hypothesize that using more fields
can improve KWS. So we create P(multi), a Polaris version that uses 3
fields: the description, the table/column name expansions
(concatenated into a string), and the table context.

P(multi) uses the following BM25 score
$\mathrm{BM25}(q, d) = \max_{f \in \mathcal{F}} \mathrm{BM25}_f(q,d)
+\; 0.1 \!\!\sum_{f' \in \mathcal{F} \setminus \{f^*\}}
\!\!\mathrm{BM25}_{f'}(q,d)$, where $\mathcal{F}$ is the set of
the above 3 fields, $f^* = \arg\max_{f} \mathrm{BM25}_f(q,d)$ is the
best-matching field, and the second term acts as a tiebreaker
(weighted by 0.1) over the remaining fields.  This ``best field plus
tiebreaker'' formulation lets the strongest signal dominate while
still rewarding tables that match across multiple fields.

Figure \ref{multifields} shows that P(multi) outperforms Polaris using
a single field on 5 datasets, often by a large margin, and is
competitive on 1 dataset. In a detailed ablation study, we found that
the table description remains the most critical field (removing it
causes significant NDCG drop). We also found that the
table/column name expansions are critical for datasets with a lot of
abbreviated table/column names (e.g., AW, LTER, Arctic).  Finally, we
found that the table context matters for Web and government data, e.g.,
removing it substantially hurts \dwtr{}, where the page-level metadata
provides important topical signals. See Appendix~\ref{sec:kws} for detail.

%% file: sections/related_work.tex
\section{Related Work}\label{sec:related}

\paragraph{Metadata Generation for Tables:} A growing body
of work has been focusing on generating metadata for tables, such as table
and dataset descriptions
\cite{autoddg,pneuma2025,singh2025rag,gan2026less,dataatlas2026},
column and database descriptions
\cite{wretblad2024synthetic,gao2025automatic,cai2026taco}, 
table or column name expansion
\cite{columbo,zhang2023nameguess,luoma2025snails,xia2025nameguess},
semantic column types
\cite{sherlock2019,sato2020,archetype2024,chorus2024}, and spatial and
temporal coverage summaries for columns \cite{auctus2021}.

     This is to target a variety of downstream tasks, including KWS,
     NL2SQL, data catalog, and more. Among these works, the closest to
     Polaris is AutoDDG \cite{autoddg}, which also generates table
     descriptions. But AutoDDG (and many prior works) does not train
     the LLM using retrieval feedback, nor uses table/column name
     expansion, as Polaris does. 

\paragraph{Keyword Search over Tables:}
Prior Keyword Search work has largely
focused on improving \emph{retrieval and ranking
models}. \citet{zhang2018adhoc} propose semantic matching between
queries and tables using sparse and dense representations within a
learning-to-rank framework. \citet{chen2020bert} train a BERT-based
neural ranker with content selection to encode tables under BERT's
input constraints, while StruBERT~\cite{trabelsi2022strubert} further
incorporates row and column structure into the matching
model. \citet{chen2020ecir} generate schema labels from table content
and incorporate the labels into a ranking model to improve dataset search.
 
In contrast, \sysname{} improves the \emph{table representation} by
generating descriptions optimized for keyword search. Our approach is
complementary to these retrieval models, since the generated
descriptions can serve as input to any of them.

Existing benchmarks for table retrieval have been discussed in
\cite{zhang2018adhoc,bhagavatula2015tabel, chen2021wtr, chen2020ecir},
providing the labeled relevance data that enables our DPO
training. More recently, TARGET~\cite{ji2025target} shows that dense
retrievers substantially outperform BM25 but remain highly sensitive
to metadata quality, while PIPER~\cite{terrenzi2026piper} addresses
poor metadata by generating pseudo-queries from table contents for
dense retrieval. Similar to \sysname{}, PIPER generates textual
representations of tables, but it targets dense retrieval and does not
optimize the generator using retrieval feedback.

\paragraph{Preference Learning and DPO:}
Reinforcement Learning from Human Feedback (RLHF)~\cite{ziegler2019fine}
aligns language models with human preferences by learning a reward model
and optimizing the policy via reinforcement learning. Direct Preference
Optimization (DPO)~\cite{dpo} simplifies this process by directly
training on preference pairs, achieving comparable performance without a
reward model. DPO has since been widely applied to tasks such as
dialogue alignment~\cite{tunstall2024zephyr} and summarization, while
RLAIF~\cite{rlaif2023} shows that AI-generated preference labels can
effectively replace human annotations.

Our work applies DPO in a different setting: preference pairs are
derived automatically from keyword retrieval performance rather than
human or AI judgments. The closest concurrent work is
\citet{santos2025dpoquery}, who optimize \emph{query} generation using
retrieval feedback. Other work similarly uses retrieval signals to
optimize query generation~\cite{zhou2023genrrl,yao2025expandr} or
document rewriting~\cite{uzan2026docopt}. In contrast, \sysname{}
optimizes \emph{table description generation}. Preference pairs are
constructed from human relevance judgments in table retrieval
benchmarks, using BM25 retrieval effectiveness as the optimization
signal.

%% file: sections/conclusion.tex
\section{Conclusion}\label{sec:conclusion}

We presented \sysname{}, a system for generating table descriptions optimized for keyword search. Our key idea is to close the loop between metadata generation and retrieval: instead of relying on human or AI judgments, we use retrieval performance on existing table retrieval benchmarks to automatically construct preference pairs and fine-tune an LLM via DPO. Combined with table and column name expansion, this produces descriptions that better match users' search vocabulary.

 Experiments on six datasets spanning enterprise, scientific, government, and web domains show that \sysname{} consistently outperforms the state of the art while using a simpler generation pipeline. More broadly, our results suggest that retrieval benchmarks can serve not only to evaluate retrieval systems, but also to train metadata generators. We hope this idea will prove useful beyond keyword search, for example in optimizing metadata for dense retrieval, NL2SQL, table question answering, and other downstream tasks.

%% file: sections/limitations.tex
\clearpage
\section{Limitations}\label{sec:limitations}

Polaris requires query--table relevance judgments to construct
preference pairs for DPO training. While several table retrieval
benchmarks already provide such labels and new ones can often be
collected from user logs or relevance annotation, our approach is less
applicable to domains where no retrieval supervision is available. In
such settings, the untrained variant of Polaris remains applicable,
although it does not benefit from retrieval-guided optimization.

Our current training objective derives preferences solely from BM25
retrieval scores. Consequently, the learned descriptions are optimized
for lexical retrieval and may not transfer to retrieval systems based
primarily on dense embeddings or hybrid ranking. Extending our
preference construction framework to optimize for dense or hybrid
retrieval objectives is an important direction for future work.

More generally, Polaris optimizes the metadata generation component
rather than the retrieval model itself. As such, it is complementary
to advances in retrieval architectures, including dense retrievers and
learned ranking models, which could potentially benefit from improved
table descriptions.

Polaris generates descriptions using only table schemas and
metadata. This design targets practical settings where table contents
are unavailable because of privacy, legal, or computational
constraints. However, when representative table rows are available,
incorporating table contents may yield richer descriptions for certain
datasets. Developing methods that selectively exploit row-level data
while preserving the simplicity and broad applicability of Polaris is
an interesting direction for future work.

Finally, our experiments cover six benchmark datasets spanning
enterprise, scientific, government, and Web domains, and
leave-one-dataset-out evaluation demonstrates encouraging cross-domain
generalization. Nevertheless, the benefit of retrieval-guided training
varies across datasets, suggesting that the effectiveness of retrieval-based
preference learning depends on characteristics of the underlying
corpus. Better understanding when retrieval feedback is most
informative remains an important direction for future work.

%% file: sections/appendix.tex
\appendix
\renewcommand{\thesection}{A.\arabic{section}}

\section{Table/Column Name Expansion}\label{sec:app-expansion}

This section provides additional detail on the name expansion step
(Section~\ref{sec:expansion}).

\twocolumn[{%
\begin{@twocolumnfalse}
  \centering
  \begin{minipage}[t]{0.29\textwidth}
    \centering
    \includegraphics[width=\textwidth]{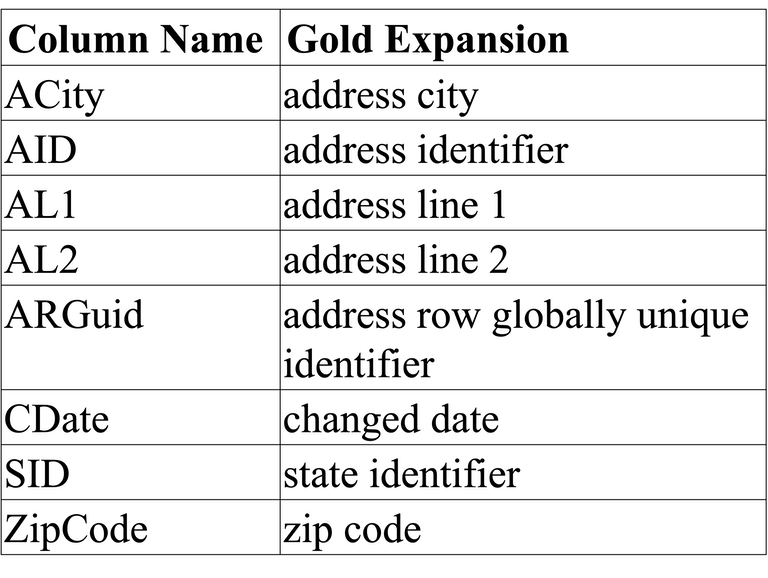}
    \smallskip\small (a) Table name: Address
  \end{minipage}
  \hfill
  \begin{minipage}[t]{0.3\textwidth}
    \centering
    \includegraphics[width=\textwidth]{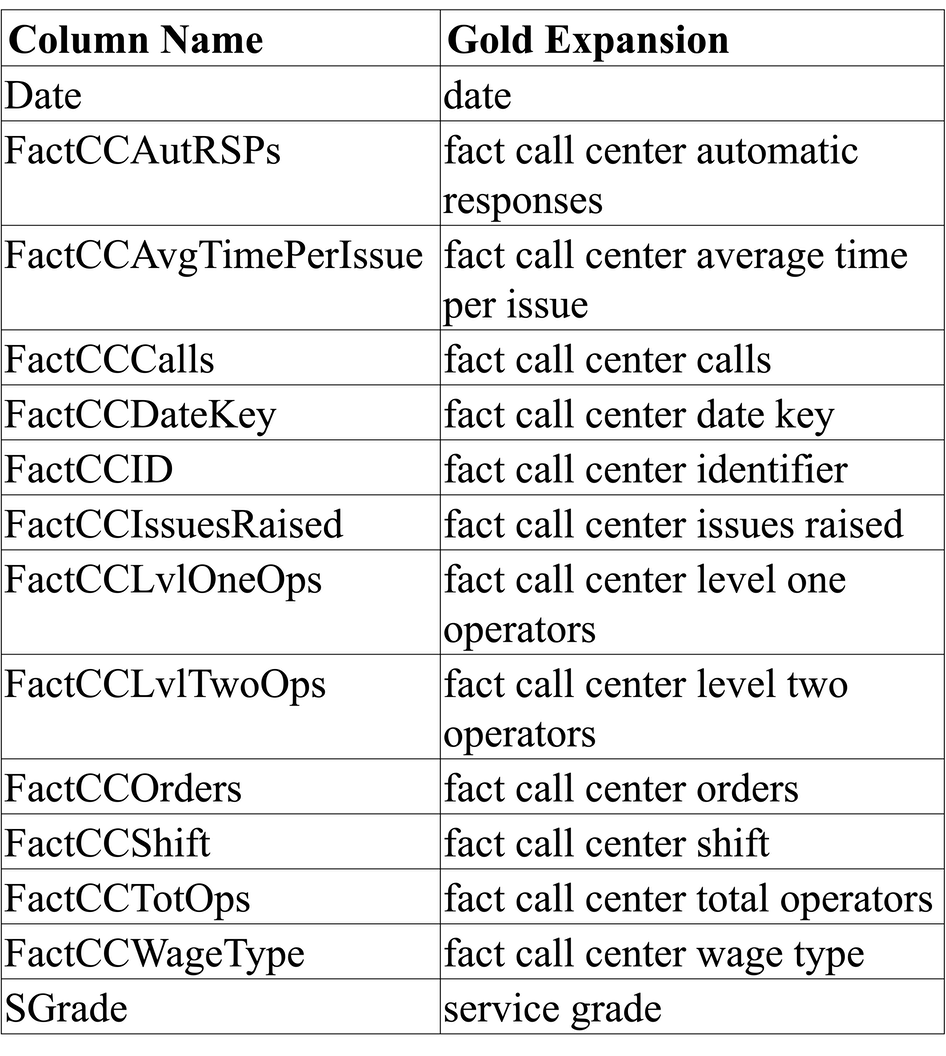}
    \smallskip\small (b) Table name: FactCallCenter
  \end{minipage}
  \hfill
  \begin{minipage}[t]{0.35\textwidth}
    \centering
    \includegraphics[width=\textwidth]{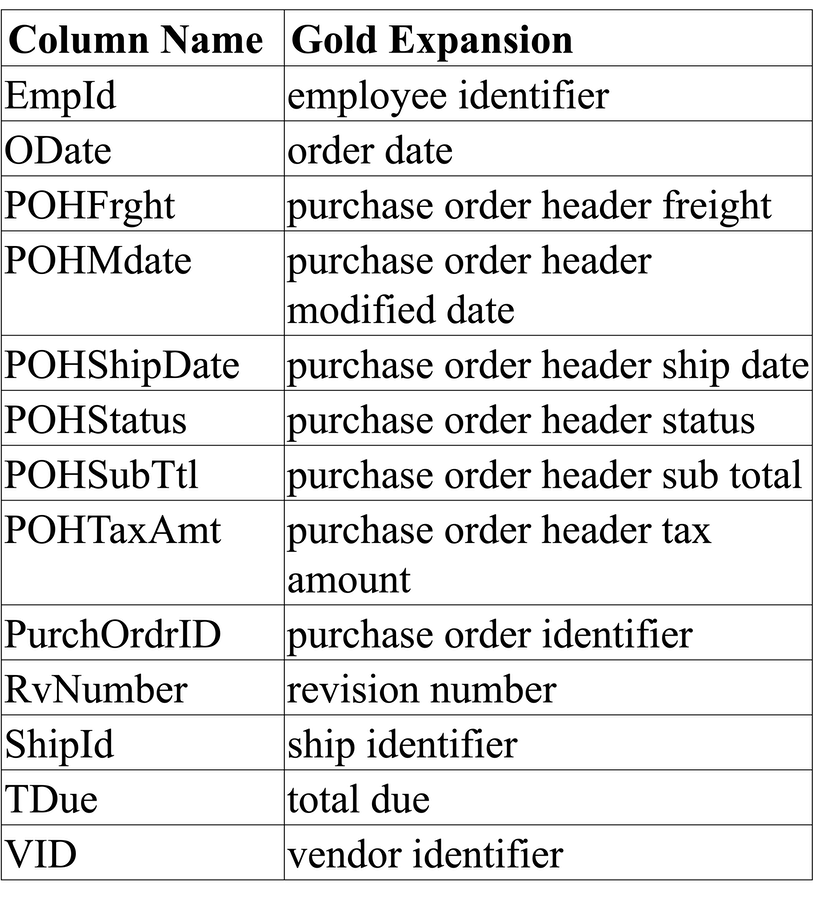}
    \smallskip\small (c) Table name: PurchaseOrderHeader
  \end{minipage}
  \captionof{figure}{Example column names and expansions.}
  \label{fig:aw-expansion}
\end{@twocolumnfalse}
}]

Figure~\ref{fig:aw-expansion} shows examples of original column
names and their gold expansions.

\begin{table*}[t!]
\centering
\begin{tcolorbox}[colback=white, colframe=black,
  title={\textbf{Column Name Expansion Prompt}}, fonttitle=\small,
  fontupper=\footnotesize, fontlower=\footnotesize]
\textbf{System:} You are a helpful assistant, answer the question from the user and reply in the same format.

\tcblower

\textbf{User:}
Your task is to expand abbreviated column names into full-form phrases.
You should return the tokens of each column name and their associated expansions.
First reason step by step and then return your final answer.
Follow the guidelines below when you expand:
\begin{enumerate}[nosep,leftmargin=*]
\item Expand all abbreviations in the column names.
\item Expand chemical symbols and units of measure to their full names.
\item Do not expand or mutate numbers.
\item Do not add extra words or explanations.
\item Maintain the original order of tokens in the expansion.
\item The tokens should be as concise and simple as possible.
\item Only provide 1 expansion for each token, even if you are uncertain, output the most possible one. Do not put ambiguous or need context after your expansion.
\item If the token is not abbreviated, the expansion should be itself, do not paraphrase or add other words around it.
\end{enumerate}

\medskip
{[Question]}\\
As abbreviations of column names from a table named Prchs\_info, c\_name | pCd | dt stand for? Think step by step\\
{[Answer]}\\
\#\#\# Reasoning\\
1. \textbf{Identify the table name context:}
\begin{itemize}[nosep,leftmargin=*]
\item The table is named \textbf{``Prchs\_info''}, which likely stands for \textbf{``Purchase Information''}.
\item This suggests that the column names are related to purchase details.
\end{itemize}
2. \textbf{Break down each column name into tokens and infer meanings:}
\begin{itemize}[nosep,leftmargin=*]
\item \textbf{c\_name} $\to$ \texttt{c}, \texttt{name}
  \begin{itemize}[nosep,leftmargin=*]
  \item \texttt{c} is commonly used for \textbf{Customer} in business contexts.
  \item \texttt{name} clearly refers to \textbf{Name}.
  \item Together, \texttt{c\_name} most likely means \textbf{Customer Name}.
  \end{itemize}
\item \textbf{pCd} $\to$ \texttt{p}, \texttt{Cd}
  \begin{itemize}[nosep,leftmargin=*]
  \item \texttt{p} is frequently used for \textbf{Product} in sales or inventory tables.
  \item \texttt{Cd} is a common abbreviation for \textbf{Code}.
  \item Together, \texttt{pCd} most likely means \textbf{Product Code}.
  \end{itemize}
\item \textbf{dt} $\to$ \texttt{dt}
  \begin{itemize}[nosep,leftmargin=*]
  \item \texttt{dt} is a standard abbreviation for \textbf{Date}.
  \item In the context of a purchase table, it likely refers to \textbf{Purchase Date}.
  \end{itemize}
\end{itemize}
\#\#\# Final Answer\\
c\_name: c $\to$ Customer, name $\to$ Name\\
pCd: p $\to$ Product, Cd $\to$ Code\\
dt: dt $\to$ Date

\medskip
\textit{(3 more few-shot examples omitted for brevity)}

\medskip
{[Question]}\\
As abbreviations of column names from a table named {\color{red}\{table\_name\}}, {\color{red}\{column\_name\_1\}} | {\color{red}\{column\_name\_2\}} | \ldots{} | {\color{red}\{column\_name\_k\}} stand for\\
{[Answer]}
\end{tcolorbox}
\captionof{table}{Column name expansion prompt. The prompt uses four few-shot examples (first shown in full above; three omitted for brevity) followed by the target input. Input placeholders are shown in {\color{red}red}.}
\label{tab:column-prompt}
\end{table*}

\paragraph{Column Name Expansion:}
Table~\ref{tab:column-prompt} shows the prompt for column name expansion. 
It has three parts.
First, a system message that fixes the assistant role
and asks the model to reply in the same format as the examples.

Second, a task statement with eight guidelines:
expand every abbreviation, including chemical symbols and units;
leave numbers untouched;
add nothing beyond the expansion;
preserve token order;
keep expansions short;
commit to exactly one expansion per token
rather than hedging with alternatives;
and leave non-abbreviated tokens unchanged.

Third, four few-shot examples
(\texttt{Prchs\_info}, \texttt{Emp\_info}, \texttt{AirQ\_data},
and \texttt{Pop\_census}),
each written as a question, a chain-of-thought reasoning block,
and a final answer.
The reasoning block first reads the table name for domain context,
then splits each column name into tokens
and expands them one at a time.

The final answer is a token-level mapping,
e.g., \texttt{sal\_amt: sal $\to$ Salary, amt $\to$ Amount},
which we concatenate into a single expanded string per column.
The target table's name and a batch of its column names
are appended in the same question format,
and the model continues from \texttt{[Answer]}.
We use the default batch size of 10 column names per prompt.

\begin{table*}[t!]
\centering
\begin{tcolorbox}[colback=white, colframe=black,
  title={\textbf{Table Name Expansion Prompt}}, fonttitle=\small,
  fontupper=\footnotesize, fontlower=\footnotesize]
\textbf{System:} You are a helpful assistant, answer the question from the user and reply in the same format.

\tcblower

\textbf{User:}
Your task is to expand abbreviated table names into full-form phrases.
You should return the tokens of each table name and their associated expansions.
First reason step by step and then return your final answer.
Follow the guidelines below when you expand:
\begin{enumerate}[nosep,leftmargin=*]
\item Expand all abbreviations in the table names.
\item Expand chemical symbols and units of measure to their full names.
\item Do not expand or mutate numbers.
\item Do not add extra words or explanations.
\item Maintain the original order of tokens in the expansion.
\item The tokens should be as concise and simple as possible.
\item Only provide 1 expansion for each token, even if you are uncertain, output the most possible one. Do not put ambiguous or need context after your expansion.
\item If the token is not abbreviated, the expansion should be itself, do not paraphrase or add other words around it.
\end{enumerate}

\medskip
{[Question]}\\
Given the following table information:\\
Table name: Prchs\_info\\
Column names: c\_name | pCd | dt

What is the full form of the table name Prchs\_info? Think step by step\\
{[Answer]}\\
\#\#\# Reasoning\\
1. \textbf{Understand the context from the table name:}
\begin{itemize}[nosep,leftmargin=*]
\item The table name is \textbf{``Prchs\_info''}.
\item The suffix \texttt{\_info} clearly suggests \textbf{``Information''}.
\item The prefix \textbf{``Prchs''} is a shortened form, likely representing \textbf{``Purchase''}.
\end{itemize}
2. \textbf{Tokenize the table name:}
\begin{itemize}[nosep,leftmargin=*]
\item Tokens: \texttt{Prchs}, \texttt{info}
\item \texttt{Prchs} $\to$ common abbreviation for \textbf{Purchase}.
\item \texttt{info} $\to$ abbreviation for \textbf{Information}.
\end{itemize}
\#\#\# Final Answer\\
Prchs\_info: Prchs $\to$ Purchase, info $\to$ Information

\medskip
\textit{(3 more few-shot examples omitted for brevity)}

\medskip
{[Question]}\\
Given the following table information:\\
Table name: {\color{red}\{table\_name\}}\\
Column names: {\color{red}\{expanded\_column\_name\_1\}} | {\color{red}\{expanded\_column\_name\_2\}} | \ldots{} | {\color{red}\{expanded\_column\_name\_k\}}

What is the full form of the table name {\color{red}\{table\_name\}}? Think step by step\\
{[Answer]}
\end{tcolorbox}
\captionof{table}{Table name expansion prompt. The prompt uses four few-shot examples (first shown in full above; three omitted for brevity) with expanded column names as additional context. Input placeholders are shown in {\color{red}red}.}
\label{tab:table-prompt}
\end{table*}

\paragraph{Table Name Expansion:}
Table~\ref{tab:table-prompt} shows the prompt for table name expansion.
It reuses the system message and the eight guidelines verbatim,
changing only ``column names'' to ``table names'' in the task statement,
and it uses the same four tables as few-shot examples.

The difference is in the input:
each question supplies the table name
\emph{together with its column names},
so the model can use the columns as evidence for what the table is about
(e.g., \texttt{hh\_id}, \texttt{age\_grp}, \texttt{edu\_lvl}
support reading \texttt{Pop\_census} as ``Population Census'').

At inference we pass the already-expanded column names
rather than the original ones,
and include at most the first 25 of them to bound prompt length.

The output is again a token-level mapping,
e.g., \texttt{Prchs\_info: Prchs $\to$ Purchase, info $\to$ Information},
concatenated into one expanded table name.
Tables without a name skip this step.

\FloatBarrier
\twocolumn[{%
\begin{@twocolumnfalse}
\section{Initial Table Description Generation}\label{sec:app-initial-generation}

\begin{minipage}[t]{0.48\textwidth}
This section provides additional detail on the initial table description
generation step: the prompt used to generate descriptions, a sample
generated description, and the LLM and decoding configuration used at
this step.
\end{minipage}
\hfill
\begin{minipage}[t]{0.48\textwidth}
\noindent\mbox{\textbf{Description Generation:}
Table~\ref{tab:desc-prompt}} shows the description generation prompt.
\end{minipage}

\medskip
\centering

\begin{tcolorbox}[colback=white, colframe=black, boxsep=1pt,
  left=2mm, right=2mm, top=1mm, bottom=1mm,
  title={\textbf{Description Generation Prompt}}, fonttitle=\small,
  fontupper=\footnotesize, fontlower=\footnotesize]
\textbf{System:}
You are a careful, concise data documentation assistant.
You ONLY use the provided table information, which may include table name,
column names and table context, to infer high-level information.
If something is not obvious from the schema, say `unknown' instead of guessing.
You do NOT need to restate the table name, column names, or table context in the JSON.
Your output MUST be valid JSON, wrapped in triple backticks with the language tag
\texttt{json}. Do NOT include any text outside the JSON code fence.
Follow the schema exactly---no extra keys.

\smallskip
\textbf{JSON schema} (instructions, not to be echoed verbatim):\\
\texttt{\{"dataset\_overview": str, "key\_themes\_or\_topics": [str],}
\texttt{"applications\_and\_use\_cases": [str],}
\texttt{"concepts\_and\_synonyms": [\{"concept": str, "synonyms": [str]\}],}
\texttt{"keywords\_and\_themes": [str], "additional\_context": [str]\}}

\smallskip
\textbf{Formatting rules:}
\begin{enumerate}[nosep,leftmargin=*]
\item Return only a JSON object inside a \texttt{```json} code fence.
\item Include no explanations, headers, or markdown outside the code fence.
\item Maintain the key order in the schema if possible.
\item For \texttt{concepts\_and\_synonyms}, include 3--6 items, each with
      1--4 synonyms.
\end{enumerate}

\tcblower

\textbf{User (one-shot example):}\\
Given the table information below:\\
Table Name: insurers\\
Column Names: company\_name, year, insurer\_type, total\_assets,
liabilities, premiums\_written\\
Produce documentation following the JSON schema in the system message.
Keep it concise and truthful. If unsure, use `unknown'.

\smallskip
\textbf{Assistant (one-shot example):}\\
\texttt{```json}\\
\texttt{\{"dataset\_overview": "Financial attributes of health insurers by company and year,}
\texttt{including insurer type, assets, liabilities, and written premiums.",}\\
\texttt{"key\_themes\_or\_topics": ["Health insurance finance", "Insurer types",}
\texttt{"Financial metrics"],}\\
\texttt{"applications\_and\_use\_cases": ["Evaluate insurer financial health",}
\texttt{"Benchmark insurer types", "Analyze market trends"],}\\
\texttt{"concepts\_and\_synonyms": [\{"concept": "Premiums Written",}
\texttt{"synonyms": ["Premium Income", "Gross Written Premium"]\},}
\texttt{\{"concept": "Liabilities", "synonyms": ["Financial Obligations", "Reserves"]\},}
\texttt{\{"concept": "Assets", "synonyms": ["Total Assets", "Asset Valuation"]\}],}\\
\texttt{"keywords\_and\_themes": ["health insurance", "premiums", "assets",}
\texttt{"liabilities", "yearly financials"],}\\
\texttt{"additional\_context": ["Values are organized by company and year."]\}}\\
\texttt{```}

\smallskip
\textbf{User (actual query):}\\
Using the table information below, generate JSON that follows the schema in
the system message. Write 2--5 sentences for \texttt{dataset\_overview}.
Provide specific, schema-derived items; avoid speculation. If information is
unclear, use `unknown'. Output JSON only, wrapped in \texttt{```json} fences.

\smallskip
Table Name: {\color{red}\{expanded\_table\_name\}} (optional)\\
Column Names: {\color{red}\{expanded\_col\_1\}},
{\color{red}\{expanded\_col\_2\}}, \ldots,
{\color{red}\{expanded\_col\_k\}}\\
Table Context: {\color{red}\{table\_context\}} (optional)
\end{tcolorbox}
\captionof{table}{Prompt for initial table-description generation. Input
placeholders are shown in {\color{red}red}.}
\label{tab:desc-prompt}
\end{@twocolumnfalse}
}]

\twocolumn[{%
\begin{@twocolumnfalse}
\begin{tcolorbox}[colback=white, colframe=black,
  title={\textbf{Sample Generated Table Description}}, fonttitle=\small,
  fontupper=\footnotesize]
\texttt{\{}\\
\quad\texttt{"dataset\_overview": "Quarterly language-preference data for Social Security}
\texttt{Disability Insurance claimants from fiscal year 2016 onward. The table supports}
\texttt{analysis of how claimant language preferences change across quarters and years.",}\\
\quad\texttt{"key\_themes\_or\_topics": ["Social Security Disability Insurance",}
\texttt{"Language preferences", "Quarterly data"],}\\
\quad\texttt{"applications\_and\_use\_cases": ["Analyze quarterly language trends",}
\texttt{"Compare claimant language preferences across fiscal years"],}\\
\quad\texttt{"concepts\_and\_synonyms": [}\\
\qquad\texttt{\{"concept": "Language Preference", "synonyms": ["Spoken Language",}
\texttt{"Preferred Language"]\},}\\
\qquad\texttt{\{"concept": "Disability Insurance", "synonyms": ["SSDI",}
\texttt{"Disability Benefits"]\},}\\
\qquad\texttt{\{"concept": "Quarterly Data", "synonyms": ["Periodic Data",}
\texttt{"Time Series"]\}],}\\
\quad\texttt{"keywords\_and\_themes": ["language preferences", "SSDI", "quarterly data",}
\texttt{"claimant demographics", "fiscal year"],}\\
\quad\texttt{"additional\_context": ["Data is organized by fiscal year and quarter."]}\\
\texttt{\}}
\end{tcolorbox}
\captionof{table}{Example output of the initial description-generation stage.}
\label{tab:app-desc-example}
\end{@twocolumnfalse}
}]

\paragraph{Sample Description:}
Table~\ref{tab:app-desc-example} shows an output description produced by the prompt.
The input schema contains fiscal-year, quarter, and language-preference
columns for Disability Insurance claimants.

\smallskip
\noindent\mbox{\textbf{The Generator:}} We use
Llama~3.1 8B Instruct as the base model.  At this initial stage,
the model has not yet been fine-tuned with DPO.  For each table, the
model receives the expanded table name, expanded
column names, and table context, and returns the
six-field description illustrated above.

For each gold table---a table relevant to at least one training query---
we sample $K=3$ descriptions with temperature~0.5,
\texttt{top\_p} = 0.9, \texttt{top\_k} = 50, and seed~42.  The non-zero
temperature gives the candidates enough variation to create a useful
ranking signal.  For every other table, we generate one description
using the same model, prompt, and decoding settings.

\FloatBarrier
\section{Ranking the Table Descriptions}
\label{sec:app-ranking}

We rank the $K$ candidate descriptions of each gold table by how well
they support retrieval for the queries related to that table.  Let
$Q(T)=\{q:R(q,T)>0\}$ be the set of queries for which table~$T$ is
relevant. We construct a
unified corpus~$C_T$ containing $K$ copies of~$T$, one with each
candidate description, and one copy of every other table.  A gold table
other than~$T$ is represented by one randomly selected candidate.  Thus,
the $K$ copies of~$T$ differ only in their descriptions.

For query~$q$ and candidate description~$d$, define the BM25
length-normalization factor

\[
\ell(d)=1-b+b\frac{|d|}{\mathrm{avgdl}}.
\]

BM25 then assigns

\begin{align*}
\mathrm{BM25}_{C_T}(q,d)
  &= \sum_{t\in q}\mathrm{IDF}(t)
     \frac{\mathrm{tf}(t,d)(k_1+1)}
     {\mathrm{tf}(t,d)+k_1\ell(d)},\\
\mathrm{IDF}(t)
  &= \log\!\left(1+\frac{N-n_t+0.5}{n_t+0.5}\right),
\end{align*}
where $\mathrm{tf}(t,d)$ is the frequency of term~$t$ in~$d$,
$|d|$ is the description length, $\mathrm{avgdl}$ is the average
document length in~$C_T$, $N=|C_T|$, and $n_t$ is the number of
documents containing~$t$.  We use Elasticsearch's BM25 defaults,
$k_1=1.2$ and $b=0.75$.

The score of candidate description~$d_j$ is its BM25 score averaged
over all queries for which~$T$ is relevant:

\[
s_T(d_j)=\frac{1}{|Q(T)|}
\sum_{q\in Q(T)}\mathrm{BM25}_{C_T}(q,d_j).
\]

This averaging gives equal weight to every relevant query and produces
one retrieval-oriented score per description.  We order the candidates
as $d_{\pi_1}\succ d_{\pi_2}\succ\cdots\succ d_{\pi_K}$, where
$s_T(d_{\pi_1})\geq s_T(d_{\pi_2})\geq\cdots\geq s_T(d_{\pi_K})$.
Intuitively, a higher-ranked description uses language that makes the
correct table score more highly for its relevant queries.
\section{DPO Training with the Rankings}
\label{sec:app-dpo}

With $K=3$, each gold table has ranked candidates
$d_{\pi_1}\succ d_{\pi_2}\succ d_{\pi_3}$.  We retain the most
separated pair and construct one training example

\[
\bigl(\mathrm{prompt}(T),\;
d_{\mathrm{win}}=d_{\pi_1},\;
d_{\mathrm{lose}}=d_{\pi_3}\bigr).
\]

Direct Preference Optimization trains the model to increase the
likelihood of the winning description relative to the losing one,
without fitting a separate reward model.  Its loss is

\begin{multline*}
\mathcal{L}_{\mathrm{DPO}}=-\mathbb{E}\!\left[\log\sigma\!\left(
\beta\log\frac{\pi_\theta(d_{\mathrm{win}}\mid x)}
{\pi_{\mathrm{ref}}(d_{\mathrm{win}}\mid x)}\right.\right.\\
\left.\left.{}-\beta\log\frac{\pi_\theta(d_{\mathrm{lose}}\mid x)}
{\pi_{\mathrm{ref}}(d_{\mathrm{lose}}\mid x)}
\right)\right],
\end{multline*}
where $x=\mathrm{prompt}(T)$, $\pi_\theta$ is the model being trained,
$\pi_{\mathrm{ref}}$ is the frozen base model, and $\beta$ controls
regularization.

We use Llama~3.1 8B Instruct with 4-bit quantization and LoRA for
parameter-efficient DPO training.  The adapter targets all attention
and feed-forward projections
(\texttt{q\_proj}, \texttt{k\_proj}, \texttt{v\_proj},
\texttt{o\_proj}, \texttt{gate\_proj}, \texttt{up\_proj}, and
\texttt{down\_proj}), with rank $r=64$ and $\alpha=64$.  This gives
167M trainable parameters out of 8.2B total (2.05\%).  We train for one
epoch with $\beta=0.1$, learning rate $5\times10^{-6}$, a linear
schedule with 10\% warmup, effective batch size~8, and the AdamW 8-bit
optimizer.  Training uses leave-one-dataset-out cross-validation: for
each held-out dataset, preference pairs from the other five datasets
form the training set.

\begin{table}[H]
\centering
\small
\begin{tabular}{ll}
\toprule
\textbf{Parameter} & \textbf{Value} \\
\midrule
Base model & Llama 3.1 8B Instruct \\
Quantization & 4-bit \\
LoRA rank / $\alpha$ & 64 / 64 \\
Trainable parameters & 167M / 8.2B (2.05\%) \\
DPO $\beta$ & 0.1 \\
Learning rate & $5\times10^{-6}$ \\
Epochs & 1 \\
Effective batch size & 8 \\
Optimizer & AdamW 8-bit \\
LR schedule & Linear; 10\% warmup \\
\bottomrule
\end{tabular}
\caption{DPO training configuration.}
\label{tab:app-dpo-config}
\end{table}

\section{Datasets and Experimental Settings}\label{sec:app-datasets}

We evaluate on six datasets spanning four domains.
The datasets vary in scale (96 to 4{,}634 tables),
domain (enterprise, scientific, government, web),
and the number of relevant tables per query.
Three datasets are curated from prior work
and three are newly created.

\paragraph{\dwiki:}
Derived from a large corpus of Wikipedia tables
used for entity linking~\cite{bhagavatula2015tabel}.
The original benchmark contains 60 queries and 1.6 million tables,
with graded relevance on a 0--2 scale
(0: irrelevant, 1: partially relevant, 2: fully relevant).
As these are web tables, they lack meaningful table names;
each table consists of column names
and table context (page title, caption, and surrounding text).

Generating descriptions for all 1.6M tables is computationally
prohibitive, so we apply a BM25-based downsampling strategy (see
below). After downsampling and removing queries with no confirmed
positive tables, we retain 3{,}361 tables and 57 queries.

\paragraph{\dwtr:}
Built over English web tables from CommonCrawl~\cite{chen2021wtr},
using the same 60 queries as \dwiki{}
but with different table coverage.
The original benchmark contains approximately 3 million tables,
also with graded relevance on a 0--2 scale.
Tables include page titles, captions,
and surrounding text as context but no table names.
We apply the same downsampling strategy,
yielding 4{,}634 tables across 60 queries.\\[2mm]
\noindent{\em Downsampling Strategy:}
Both \dwiki{} and \dwtr{} contain millions of tables,
making it impractical to generate descriptions for all of them.
We downsample each corpus deterministically
using a BM25-based strategy.
For each query, we retrieve the top-50 tables
from the full corpus via BM25,
using only the available metadata fields
(column names with the trigram analyzer
and table context with the word analyzer,
both with equal weight);
no descriptions or table names are used,
as descriptions have not yet been generated at this stage
and these datasets do not have table names.
We then take the union of these retrieved tables
with all gold-relevant tables (relevance score $> 0$).
This strategy does not reduce task difficulty;
rather, it focuses the corpus on hard cases
by retaining the most competitive non-relevant tables
(hard negatives that BM25 would actually surface)
alongside all truly relevant tables.

\paragraph{\decir:}
Built from the U.S.\ government open data portal~\cite{chen2020ecir}.
The original benchmark defines 6 search tasks,
each with 20 queries,
and provides task--table relevance judgments
on a 0--3 scale from multiple annotators.
Prior work~\cite{autoddg} treats all 120 queries independently.
Upon careful inspection, we identify two issues:
(1)~the 20 queries within each task are highly similar
and lack diversity, with some queries
unrelated to the task; and
(2)~the relevance labels suffer from false positives---many
table--task pairs labeled as highly relevant ($\geq 2$)
are not actually relevant upon manual inspection,
while pairs labeled as irrelevant (0) are generally correct.

To address these issues, we manually re-label all table--query pairs
originally scored as highly relevant ($\geq 2$), keeping labels of 0
as-is.  To reduce redundancy across queries within a task, we select 2
representative and maximally distinct queries per task.  We also
remove tables that are not in CSV format and cannot be opened,
yielding 12 queries across 2{,}100 parseable tables with relevance on
a 0--3 scale.

\paragraph{\daw:}
Based on the AdventureWorks sample database (Microsoft),
a well-known enterprise schema with 96 tables
that use many abbreviated names
(e.g., \texttt{BusEntId}, \texttt{DTPh}, \texttt{STRGUID}).
We created 15 queries for enterprise data discovery tasks,
such as ``employee phone number'' and ``discounted products'',
and manually labeled the relevance of all 96 tables for each query.
The relevance score is binary (0 or 1).

\paragraph{\darctic:}
A random sample of 251 tables from the
Environmental Data Initiative (EDI),
a scientific data repository
storing long-term ecological and environmental research data.
We created 20 queries covering topics such as
``precipitation,'' ``species diversity,'' and ``nitrous oxide''
and manually labeled the table--query pairs.
Relevance is binary (0 or 1).

\paragraph{\dlter:}
A larger sample of 2{,}015 tables
from the top-downloaded EDI collections,
representing a more focused subset
of frequently used scientific datasets.
We created 15 queries and manually labeled relevance as in \darctic{}.
Relevance is binary (0 or 1).

\paragraph{Relevance Annotation:}
For all three new datasets,
we use a pooling-based annotation workflow:
for each query, we run multiple retrieval configurations
(varying BM25 field weights, tokenization strategies,
and dense retrieval)
and collect the union of candidate tables.
A domain expert then manually verifies
whether each candidate is relevant to the query,
producing binary relevance labels
used for both training and evaluation.
This pooling approach reduces the chance of missing
relevant tables that any single retrieval method would overlook.

\paragraph{The Generators:}
We use Llama 3.1 8B Instruct~\red{\cite{llama3}} as the base model for description generation,
with 4-bit quantization (QLoRA\red{~\cite{qlora}}) for parameter-efficient DPO fine-tuning.

For name expansion, we use Llama 4 Maverick 17B Instruct via AWS Bedrock
with temperature~0.2, top\_p~0.9, and max\_tokens~4196.

Table~\ref{tab:app-dpo-config} summarizes the DPO training hyperparameters.
The LoRA adapter targets all attention and feed-forward projections
(\texttt{q\_proj}, \texttt{k\_proj}, \texttt{v\_proj}, \texttt{o\_proj},
\texttt{gate\_proj}, \texttt{up\_proj}, \texttt{down\_proj}),
with rank $r = 64$ and $\alpha = 64$,
resulting in 167M trainable parameters out of 8.2B total (2.05\%).

We train for one epoch with DPO $\beta = 0.1$, learning rate $5 \times
10^{-6}$ with linear schedule and 10\% warmup, effective batch size~8,
and AdamW 8-bit optimizer.  For each held-out dataset, we train on
preference pairs from the remaining five datasets
(leave-one-dataset-out cross-validation).

\paragraph{Keyword Search:}
We use Elasticsearch with BM25 scoring ($k_1 = 1.2$, $b = 0.75$).
The word-level analyzer applies standard tokenization
(splitting on whitespace and punctuation),
lowercasing, and English stemming (Porter),
so that query and document terms are matched
after normalization.
Unless otherwise noted (\S\ref{sec:kws}),
we index each table as a \emph{single description field}
with this analyzer,
isolating the effect of description quality
from indexing configuration differences.

\paragraph{Evaluation Metrics:}
We report NDCG@$K$ (ranking quality with graded relevance)
and Recall@$K$ (fraction of relevant tables retrieved in the top $K$)
for $K \in \{5, 10, 20, 50, 100\}$.
For Recall, all tables with relevance score $> 0$ are treated as relevant.

\section{Overall Performance}\label{sec:overall}

\begin{figure}[t]
\centering
\includegraphics[width=\columnwidth]{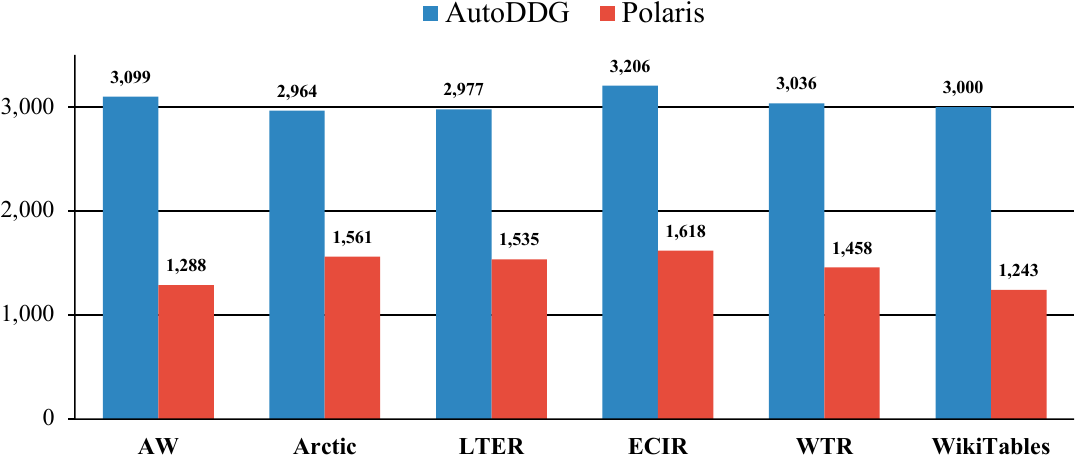}
\caption{Average length of the descriptions generated by AutoDDG and \sysname{} on each dataset.}
\label{desclen}
\end{figure}

Figure ~\ref{desclen} shows the average length of the descriptions generated by AutoDDG and Polaris.

\section{Ablation Studies}\label{sec:app-ablation}

Table~\ref{tab:prompt-ndcg} shows NDCG@K and Table~\ref{tab:prompt-recall}
shows Recall@K for prompt quality evaluation. 

\begin{table}[t]
\centering
\footnotesize
\resizebox{\columnwidth}{!}{%
\begin{tabular}{llccccc}
\toprule
Dataset & Method & @5 & @10 & @20 & @50 & @100 \\
\midrule
\daw{} & AutoDDG & .5551 & .6192 & .6692 & .6969 & .7030 \\
       & $P{-}(\mathrm{exp{+}DPO})$ & \textcolor{blue}{.7667} & \textcolor{blue}{.7948} & \textcolor{blue}{.8129} & \textcolor{blue}{.8129} & \textcolor{blue}{.8129} \\
\dsep
\darctic{} & AutoDDG & .6274 & .5942 & .5942 & .6360 & .6812 \\
           & $P{-}(\mathrm{exp{+}DPO})$ & \textcolor{blue}{.7041} & \textcolor{blue}{.6651} & \textcolor{blue}{.6700} & \textcolor{blue}{.7008} & \textcolor{blue}{.7226} \\
\dsep
\dlter{} & AutoDDG & .4372 & .4194 & .4011 & .3656 & .3559 \\
         & $P{-}(\mathrm{exp{+}DPO})$ & \textcolor{blue}{.6042} & \textcolor{blue}{.5856} & \textcolor{blue}{.5353} & \textcolor{blue}{.4358} & \textcolor{blue}{.4188} \\
\dsep
\decir{} & AutoDDG & .4215 & .4429 & .4512 & .4408 & .4180 \\
         & $P{-}(\mathrm{exp{+}DPO})$ & \textcolor{blue}{.5390} & \textcolor{blue}{.5555} & \textcolor{blue}{.5385} & \textcolor{blue}{.5141} & \textcolor{blue}{.4927} \\
\dsep
\dwtr{} & AutoDDG & \textcolor{blue}{.3748} & .3548 & \textcolor{blue}{.3767} & .4159 & .4884 \\
        & $P{-}(\mathrm{exp{+}DPO})$ & .3693 & \textcolor{blue}{.3688} & .3684 & \textcolor{blue}{.4335} & \textcolor{blue}{.5155} \\
\dsep
\dwiki{} & AutoDDG & \textcolor{blue}{.4139} & \textcolor{blue}{.4085} & .4398 & .5176 & \textcolor{blue}{.5811} \\
         & $P{-}(\mathrm{exp{+}DPO})$ & .3999 & .4027 & \textcolor{blue}{.4456} & \textcolor{blue}{.5189} & .5721 \\
\bottomrule
\end{tabular}%
}
\caption{Prompt quality ablation (NDCG@$K$): AutoDDG vs.\
$P{-}(\mathrm{exp{+}DPO})$.
$P{-}(\mathrm{exp{+}DPO})$ = \sysname{} without DPO training
and table/column name expansion.
Blue = better per cell.}
\label{tab:prompt-ndcg}
\end{table}

\begin{table}[t]
\centering
\footnotesize
\resizebox{\columnwidth}{!}{%
\begin{tabular}{llccccc}
\toprule
Dataset & Method & @5 & @10 & @20 & @50 & @100 \\
\midrule
\daw{} & AutoDDG & .49 & .64 & .76 & \textcolor{blue}{.83} & \textcolor{blue}{.85} \\
       & $P{-}(\mathrm{exp{+}DPO})$ & \textcolor{blue}{.69} & \textcolor{blue}{.78} & .82 & .82 & .82 \\
\dsep
\darctic{} & AutoDDG & .24 & .34 & .48 & .68 & \textcolor{blue}{.77} \\
           & $P{-}(\mathrm{exp{+}DPO})$ & \textcolor{blue}{.29} & \textcolor{blue}{.40} & \textcolor{blue}{.55} & \textcolor{blue}{.74} & .78 \\
\dsep
\dlter{} & AutoDDG & .08 & .14 & .22 & .32 & .35 \\
         & $P{-}(\mathrm{exp{+}DPO})$ & \textcolor{blue}{.12} & \textcolor{blue}{.18} & \textcolor{blue}{.26} & \textcolor{blue}{.33} & \textcolor{blue}{.37} \\
\dsep
\decir{} & AutoDDG & .01 & .02 & .04 & .08 & .12 \\
         & $P{-}(\mathrm{exp{+}DPO})$ & \textcolor{blue}{.02} & .02 & .04 & \textcolor{blue}{.09} & \textcolor{blue}{.14} \\
\dsep
\dwtr{} & AutoDDG & \textcolor{blue}{.08} & .14 & \textcolor{blue}{.25} & .43 & .59 \\
        & $P{-}(\mathrm{exp{+}DPO})$ & \textcolor{blue}{.08} & .14 & .24 & \textcolor{blue}{.47} & \textcolor{blue}{.64} \\
\dsep
\dwiki{} & AutoDDG & \textcolor{blue}{.21} & \textcolor{blue}{.29} & .41 & .66 & \textcolor{blue}{.83} \\
         & $P{-}(\mathrm{exp{+}DPO})$ & .20 & .28 & \textcolor{blue}{.44} & \textcolor{blue}{.68} & .82 \\
\bottomrule
\end{tabular}%
}
\caption{Prompt quality ablation (Recall@$K$): AutoDDG vs.\
$P{-}(\mathrm{exp{+}DPO})$.
$P{-}(\mathrm{exp{+}DPO})$ = \sysname{} without DPO training
and table/column name expansion.
Blue = better per cell.}
\label{tab:prompt-recall}
\end{table}

Table~\ref{tab:ablation-ndcg} reports NDCG@K and 
Table~\ref{tab:ablation-recall} reports Recall@K for the effect of DPO training and the
effect of name expansion. 

\begin{table}[t]
\centering
\footnotesize
\resizebox{\columnwidth}{!}{%
\begin{tabular}{llccccc}
\toprule
Dataset & Method & @5 & @10 & @20 & @50 & @100 \\
\midrule
\daw{} & \sysname{}     & \textcolor{blue}{.7312} & \textcolor{blue}{.7582} & \textcolor{blue}{.7869} & \textcolor{blue}{.7869} & \textcolor{blue}{.7869} \\
       & $P{-}\mathrm{DPO}$  & .6681 & .7287 & .7794 & .7794 & .7794 \\
       & $P{-}\mathrm{exp}$    & .6940 & .7485 & .7847 & .7847 & .7847 \\
\dsep
\darctic{} & \sysname{}     & .7133 & .6743 & \textcolor{blue}{.6939} & \textcolor{blue}{.7416} & \textcolor{blue}{.7678} \\
           & $P{-}\mathrm{DPO}$  & \textcolor{blue}{.7209} & \textcolor{blue}{.6785} & .6891 & .7334 & .7535 \\
           & $P{-}\mathrm{exp}$    & .6690 & .6493 & .6737 & .6872 & .7106 \\
\dsep
\dlter{} & \sysname{}     & \textcolor{blue}{.7218} & \textcolor{blue}{.6546} & \textcolor{blue}{.6013} & \textcolor{blue}{.5239} & \textcolor{blue}{.5145} \\
         & $P{-}\mathrm{DPO}$  & .6538 & .6187 & .5639 & .4743 & .4488 \\
         & $P{-}\mathrm{exp}$    & .5957 & .5750 & .5257 & .4434 & .4230 \\
\dsep
\decir{} & \sysname{}     & .4946 & .5050 & .4981 & .4873 & .4672 \\
         & $P{-}\mathrm{DPO}$  & \textcolor{blue}{.5696} & \textcolor{blue}{.5894} & \textcolor{blue}{.5502} & \textcolor{blue}{.5114} & \textcolor{blue}{.4941} \\
         & $P{-}\mathrm{exp}$    & .4912 & .4968 & .4954 & .5048 & .4726 \\
\dsep
\dwtr{} & \sysname{}     & \textcolor{blue}{.3840} & \textcolor{blue}{.3840} & \textcolor{blue}{.3894} & \textcolor{blue}{.4449} & \textcolor{blue}{.5309} \\
        & $P{-}\mathrm{DPO}$  & .3657 & .3635 & .3740 & .4402 & .5226 \\
        & $P{-}\mathrm{exp}$    & .3554 & .3781 & .3820 & .4390 & .5208 \\
\dsep
\dwiki{} & \sysname{}     & .4111 & \textcolor{blue}{.4215} & \textcolor{blue}{.4407} & .5171 & \textcolor{blue}{.5727} \\
         & $P{-}\mathrm{DPO}$  & \textcolor{blue}{.4162} & .4094 & .4396 & \textcolor{blue}{.5229} & .5722 \\
         & $P{-}\mathrm{exp}$    & .3818 & .3930 & .4377 & .5200 & .5725 \\
\bottomrule
\end{tabular}%
}
\caption{Effect of DPO training and table/column name expansion
(NDCG@$K$): \sysname{} vs.\ $P{-}\mathrm{DPO}$ vs.\ $P{-}\mathrm{exp}$.
$P{-}\mathrm{DPO}$ = \sysname{} without DPO training,
$P{-}\mathrm{exp}$ = \sysname{} without name expansion.
Blue = best per cell.}
\label{tab:ablation-ndcg}
\end{table}

\begin{table}[t]
\centering
\footnotesize
\resizebox{\columnwidth}{!}{%
\begin{tabular}{llccccc}
\toprule
Dataset & Method & @5 & @10 & @20 & @50 & @100 \\
\midrule
\daw{} & \sysname{}     & \textcolor{blue}{.66} & .77 & .83 & .83 & .83 \\
       & $P{-}\mathrm{DPO}$  & .61 & .76 & \textcolor{blue}{.89} & \textcolor{blue}{.89} & \textcolor{blue}{.89} \\
       & $P{-}\mathrm{exp}$    & .65 & \textcolor{blue}{.79} & .87 & .87 & .87 \\
\dsep
\darctic{} & \sysname{}     & .29 & \textcolor{blue}{.41} & \textcolor{blue}{.58} & \textcolor{blue}{.77} & \textcolor{blue}{.82} \\
           & $P{-}\mathrm{DPO}$  & \textcolor{blue}{.30} & .40 & .56 & .76 & .80 \\
           & $P{-}\mathrm{exp}$    & .27 & .39 & .57 & .72 & .77 \\
\dsep
\dlter{} & \sysname{}     & \textcolor{blue}{.18} & \textcolor{blue}{.25} & \textcolor{blue}{.32} & \textcolor{blue}{.39} & \textcolor{blue}{.43} \\
         & $P{-}\mathrm{DPO}$  & .12 & .19 & .28 & .36 & .38 \\
         & $P{-}\mathrm{exp}$    & .13 & .18 & .25 & .33 & .36 \\
\dsep
\decir{} & \sysname{}     & .01 & \textcolor{blue}{.03} & .04 & .08 & .13 \\
         & $P{-}\mathrm{DPO}$  & .01 & \textcolor{blue}{.03} & .04 & \textcolor{blue}{.09} & \textcolor{blue}{.14} \\
         & $P{-}\mathrm{exp}$    & .01 & .02 & .04 & \textcolor{blue}{.09} & .13 \\
\dsep
\dwtr{} & \sysname{}     & .08 & \textcolor{blue}{.15} & \textcolor{blue}{.26} & \textcolor{blue}{.48} & \textcolor{blue}{.66} \\
        & $P{-}\mathrm{DPO}$  & .08 & .14 & .25 & .47 & .65 \\
        & $P{-}\mathrm{exp}$    & .08 & \textcolor{blue}{.15} & \textcolor{blue}{.26} & \textcolor{blue}{.48} & .65 \\
\dsep
\dwiki{} & \sysname{}     & \textcolor{blue}{.21} & \textcolor{blue}{.32} & .42 & .67 & .81 \\
         & $P{-}\mathrm{DPO}$  & .20 & .28 & .41 & \textcolor{blue}{.68} & .81 \\
         & $P{-}\mathrm{exp}$    & \textcolor{blue}{.21} & .29 & \textcolor{blue}{.45} & \textcolor{blue}{.70} & \textcolor{blue}{.86} \\
\bottomrule
\end{tabular}%
}
\caption{Effect of DPO training and table/column name expansion
(Recall@$K$): \sysname{} vs.\ $P{-}\mathrm{DPO}$ vs.\ $P{-}\mathrm{exp}$.
$P{-}\mathrm{DPO}$ = \sysname{} without DPO training,
$P{-}\mathrm{exp}$ = \sysname{} without name expansion.
Blue = best per cell.}
\label{tab:ablation-recall}
\end{table}

\section{Sensitivity Analysis with Different LLMs}\label{sec:app-sens-llm}

Table~\ref{tab:sens-llm-polaris} shows NDCG@K and
Table~\ref{tab:sens-llm-recall} shows Recall@K for the effect of
replacing the underlying LLM with Qwen2.5-7B-Instruct~\cite{qwen25}.

Table~\ref{tab:sens-llm-vs} reports NDCG@K and
Table~\ref{tab:sens-llm-vs-recall} reports Recall@K for the comparison
between \sysname{} and AutoDDG, both using Qwen2.5-7B-Instruct.

\begin{table}[t]
\centering
\footnotesize
\resizebox{\columnwidth}{!}{%
\begin{tabular}{llccccc}
\toprule
Dataset & Method & @5 & @10 & @20 & @50 & @100 \\
\midrule
\daw{} & $P(\mathrm{Qwen})$  & \textcolor{blue}{.8265} & \textcolor{blue}{.8390} & \textcolor{blue}{.8733} & \textcolor{blue}{.8733} & \textcolor{blue}{.8733} \\
       & \sysname{} & .7312 & .7582 & .7869 & .7869 & .7869 \\
\dsep
\darctic{} & $P(\mathrm{Qwen})$  & \textcolor{blue}{.7303} & \textcolor{blue}{.6952} & \textcolor{blue}{.7071} & \textcolor{blue}{.7425} & \textcolor{blue}{.7690} \\
           & \sysname{} & .7133 & .6743 & .6939 & .7416 & .7678 \\
\dsep
\dlter{} & $P(\mathrm{Qwen})$  & .6269 & .5964 & .5393 & .4439 & .4259 \\
         & \sysname{} & \textcolor{blue}{.7218} & \textcolor{blue}{.6546} & \textcolor{blue}{.6013} & \textcolor{blue}{.5239} & \textcolor{blue}{.5145} \\
\dsep
\decir{} & $P(\mathrm{Qwen})$  & \textcolor{blue}{.5188} & \textcolor{blue}{.5117} & .4935 & .4834 & .4577 \\
         & \sysname{} & .4946 & .5050 & \textcolor{blue}{.4981} & \textcolor{blue}{.4873} & \textcolor{blue}{.4672} \\
\dsep
\dwtr{} & $P(\mathrm{Qwen})$  & \textcolor{blue}{.4346} & \textcolor{blue}{.3961} & \textcolor{blue}{.3957} & .4430 & .5123 \\
        & \sysname{} & .3840 & .3840 & .3894 & \textcolor{blue}{.4449} & \textcolor{blue}{.5309} \\
\dsep
\dwiki{} & $P(\mathrm{Qwen})$  & \textcolor{blue}{.4250} & \textcolor{blue}{.4383} & \textcolor{blue}{.4692} & \textcolor{blue}{.5403} & \textcolor{blue}{.5971} \\
         & \sysname{} & .4111 & .4215 & .4407 & .5171 & .5727 \\
\bottomrule
\end{tabular}%
}
\caption{\sysname{} with different LLMs (NDCG@$K$):
$P(\mathrm{Qwen})$ vs.\ \sysname{}.
$P(\mathrm{Qwen})$ = \sysname{} with Qwen2.5-7B,
\sysname{} = \sysname{} with Llama 3.1 8B.
Blue = better per cell.}
\label{tab:sens-llm-polaris}
\end{table}

\begin{table}[t]
\centering
\footnotesize
\resizebox{\columnwidth}{!}{%
\begin{tabular}{llccccc}
\toprule
Dataset & Method & @5 & @10 & @20 & @50 & @100 \\
\midrule
\daw{} & $P(\mathrm{Qwen})$  & \textcolor{blue}{.73} & \textcolor{blue}{.81} & \textcolor{blue}{.88} & \textcolor{blue}{.88} & \textcolor{blue}{.88} \\
       & \sysname{} & .66 & .77 & .83 & .83 & .83 \\
\dsep
\darctic{} & $P(\mathrm{Qwen})$  & .29 & \textcolor{blue}{.42} & \textcolor{blue}{.60} & .77 & .82 \\
           & \sysname{} & .29 & .41 & .58 & .77 & .82 \\
\dsep
\dlter{} & $P(\mathrm{Qwen})$  & .11 & .18 & .27 & .34 & .36 \\
         & \sysname{} & \textcolor{blue}{.18} & \textcolor{blue}{.25} & \textcolor{blue}{.32} & \textcolor{blue}{.39} & \textcolor{blue}{.43} \\
\dsep
\decir{} & $P(\mathrm{Qwen})$  & .01 & .02 & .04 & .08 & .12 \\
         & \sysname{} & .01 & \textcolor{blue}{.03} & .04 & .08 & \textcolor{blue}{.13} \\
\dsep
\dwtr{} & $P(\mathrm{Qwen})$  & \textcolor{blue}{.09} & .15 & .25 & .46 & .61 \\
        & \sysname{} & .08 & \textcolor{blue}{.15} & \textcolor{blue}{.26} & \textcolor{blue}{.48} & \textcolor{blue}{.66} \\
\dsep
\dwiki{} & $P(\mathrm{Qwen})$  & .19 & .30 & \textcolor{blue}{.45} & \textcolor{blue}{.69} & \textcolor{blue}{.84} \\
         & \sysname{} & \textcolor{blue}{.21} & \textcolor{blue}{.32} & .42 & .67 & .81 \\
\bottomrule
\end{tabular}%
}
\caption{\sysname{} with different LLMs (Recall@$K$):
$P(\mathrm{Qwen})$ vs.\ \sysname{}.
$P(\mathrm{Qwen})$ = \sysname{} with Qwen2.5-7B,
\sysname{} = \sysname{} with Llama 3.1 8B.
Blue = better per cell.}
\label{tab:sens-llm-recall}
\end{table}

\begin{table}[t]
\centering
\footnotesize
\resizebox{\columnwidth}{!}{%
\begin{tabular}{llccccc}
\toprule
Dataset & Method & @5 & @10 & @20 & @50 & @100 \\
\midrule
\daw{} & $A(\mathrm{Qwen})$ & .4853 & .5286 & .5946 & .6306 & .6306 \\
       & $P(\mathrm{Qwen})$  & \textcolor{blue}{.8265} & \textcolor{blue}{.8390} & \textcolor{blue}{.8733} & \textcolor{blue}{.8733} & \textcolor{blue}{.8733} \\
\dsep
\darctic{} & $A(\mathrm{Qwen})$ & .6861 & .6586 & .6786 & .7055 & .7437 \\
           & $P(\mathrm{Qwen})$  & \textcolor{blue}{.7303} & \textcolor{blue}{.6952} & \textcolor{blue}{.7071} & \textcolor{blue}{.7425} & \textcolor{blue}{.7690} \\
\dsep
\dlter{} & $A(\mathrm{Qwen})$ & .3379 & .3269 & .3275 & .3380 & .3440 \\
         & $P(\mathrm{Qwen})$  & \textcolor{blue}{.6269} & \textcolor{blue}{.5964} & \textcolor{blue}{.5393} & \textcolor{blue}{.4439} & \textcolor{blue}{.4259} \\
\dsep
\decir{} & $A(\mathrm{Qwen})$ & .3660 & .4175 & .4200 & .4033 & .3989 \\
         & $P(\mathrm{Qwen})$  & \textcolor{blue}{.5188} & \textcolor{blue}{.5117} & \textcolor{blue}{.4935} & \textcolor{blue}{.4834} & \textcolor{blue}{.4577} \\
\dsep
\dwtr{} & $A(\mathrm{Qwen})$ & .3541 & .3714 & .3691 & .4164 & .5010 \\
        & $P(\mathrm{Qwen})$  & \textcolor{blue}{.4346} & \textcolor{blue}{.3961} & \textcolor{blue}{.3957} & \textcolor{blue}{.4430} & \textcolor{blue}{.5123} \\
\dsep
\dwiki{} & $A(\mathrm{Qwen})$ & \textcolor{blue}{.4414} & \textcolor{blue}{.4493} & \textcolor{blue}{.4718} & \textcolor{blue}{.5479} & \textcolor{blue}{.6036} \\
         & $P(\mathrm{Qwen})$  & .4250 & .4383 & .4692 & .5403 & .5971 \\
\bottomrule
\end{tabular}%
}
\caption{\sysname{} vs.\ AutoDDG under Qwen2.5-7B (NDCG@$K$):
$A(\mathrm{Qwen})$ vs.\ $P(\mathrm{Qwen})$.
$A(\mathrm{Qwen})$ = AutoDDG with Qwen2.5-7B,
$P(\mathrm{Qwen})$ = \sysname{} with Qwen2.5-7B.
Blue = better per cell.}
\label{tab:sens-llm-vs}
\end{table}

\begin{table}[t]
\centering
\footnotesize
\resizebox{\columnwidth}{!}{%
\begin{tabular}{llccccc}
\toprule
Dataset & Method & @5 & @10 & @20 & @50 & @100 \\
\midrule
\daw{} & $A(\mathrm{Qwen})$ & .42 & .55 & .73 & .83 & .83 \\
       & AutoDDG & .49 & .64 & .76 & .83 & \textcolor{blue}{.85} \\
       & $P(\mathrm{Qwen})$  & \textcolor{blue}{.73} & \textcolor{blue}{.81} & \textcolor{blue}{.88} & \textcolor{blue}{.88} & \textcolor{blue}{.88} \\
       & \sysname{} & .66 & .77 & .83 & .83 & .83 \\
\dsep
\darctic{} & $A(\mathrm{Qwen})$ & .29 & .41 & .58 & .74 & .81 \\
           & AutoDDG & .24 & .34 & .48 & .68 & .77 \\
           & $P(\mathrm{Qwen})$  & .29 & \textcolor{blue}{.42} & \textcolor{blue}{.60} & \textcolor{blue}{.77} & \textcolor{blue}{.82} \\
           & \sysname{} & \textcolor{blue}{.29} & .41 & .58 & \textcolor{blue}{.77} & \textcolor{blue}{.82} \\
\dsep
\dlter{} & $A(\mathrm{Qwen})$ & .05 & .08 & .17 & .31 & .36 \\
         & AutoDDG & .08 & .14 & .22 & .32 & .35 \\
         & $P(\mathrm{Qwen})$  & .11 & .18 & .27 & .34 & .36 \\
         & \sysname{} & \textcolor{blue}{.18} & \textcolor{blue}{.25} & \textcolor{blue}{.32} & \textcolor{blue}{.39} & \textcolor{blue}{.43} \\
\dsep
\decir{} & $A(\mathrm{Qwen})$ & .01 & .02 & .04 & .07 & .11 \\
         & AutoDDG & .01 & .02 & .04 & .08 & .12 \\
         & $P(\mathrm{Qwen})$  & .01 & .02 & .04 & .08 & .12 \\
         & \sysname{} & .01 & \textcolor{blue}{.03} & .04 & .08 & \textcolor{blue}{.13} \\
\dsep
\dwtr{} & $A(\mathrm{Qwen})$ & .08 & \textcolor{blue}{.15} & .24 & .44 & .62 \\
        & AutoDDG & .08 & .14 & .25 & .43 & .59 \\
        & $P(\mathrm{Qwen})$  & \textcolor{blue}{.09} & \textcolor{blue}{.15} & .25 & .46 & .61 \\
        & \sysname{} & .08 & \textcolor{blue}{.15} & \textcolor{blue}{.26} & \textcolor{blue}{.48} & \textcolor{blue}{.66} \\
\dsep
\dwiki{} & $A(\mathrm{Qwen})$ & \textcolor{blue}{.24} & \textcolor{blue}{.34} & \textcolor{blue}{.47} & \textcolor{blue}{.70} & \textcolor{blue}{.86} \\
         & AutoDDG & .21 & .29 & .41 & .66 & .83 \\
         & $P(\mathrm{Qwen})$  & .19 & .30 & .45 & .69 & .84 \\
         & \sysname{} & .21 & .32 & .42 & .67 & .81 \\
\bottomrule
\end{tabular}%
}
\caption{\sysname{} vs.\ AutoDDG under different LLMs (Recall@$K$):
$A(\mathrm{Qwen})$ vs.\ AutoDDG vs.\ $P(\mathrm{Qwen})$ vs.\ \sysname{}.
$A(\mathrm{Qwen})$ = AutoDDG with Qwen2.5-7B,
$P(\mathrm{Qwen})$ = \sysname{} with Qwen2.5-7B;
AutoDDG and \sysname{} use Llama 3.1 8B.
Blue = best per cell.}
\label{tab:sens-llm-vs-recall}
\end{table}

\section{Sensitivity Analysis with Different DPO $\beta$}
\label{sec:app-sens-beta}

Table~\ref{tab:sens-beta} shows NDCG@K and
Table~\ref{tab:sens-beta-recall} shows Recall@K for the effect of
increasing the DPO regularization parameter $\beta$ from 0.1 to 0.5.

\begin{table}[t]
\centering
\footnotesize
\resizebox{\columnwidth}{!}{%
\begin{tabular}{llccccc}
\toprule
Dataset & Method & @5 & @10 & @20 & @50 & @100 \\
\midrule
\daw{} & $P(\beta{=}0.5)$  & \textcolor{blue}{.7398} & \textcolor{blue}{.7662} & \textcolor{blue}{.8051} & \textcolor{blue}{.8051} & \textcolor{blue}{.8051} \\
       & \sysname{} & .7312 & .7582 & .7869 & .7869 & .7869 \\
\dsep
\darctic{} & $P(\beta{=}0.5)$  & \textcolor{blue}{.7266} & \textcolor{blue}{.6874} & .6837 & .7380 & .7585 \\
           & \sysname{} & .7133 & .6743 & \textcolor{blue}{.6939} & \textcolor{blue}{.7416} & \textcolor{blue}{.7678} \\
\dsep
\dlter{} & $P(\beta{=}0.5)$  & .6760 & .6151 & .5806 & .4929 & .4732 \\
         & \sysname{} & \textcolor{blue}{.7218} & \textcolor{blue}{.6546} & \textcolor{blue}{.6013} & \textcolor{blue}{.5239} & \textcolor{blue}{.5145} \\
\dsep
\decir{} & $P(\beta{=}0.5)$  & .4914 & \textcolor{blue}{.5158} & .4837 & \textcolor{blue}{.4956} & \textcolor{blue}{.4784} \\
         & \sysname{} & \textcolor{blue}{.4946} & .5050 & \textcolor{blue}{.4981} & .4873 & .4672 \\
\dsep
\dwtr{} & $P(\beta{=}0.5)$  & .3728 & .3820 & .3868 & \textcolor{blue}{.4455} & .5275 \\
        & \sysname{} & \textcolor{blue}{.3840} & \textcolor{blue}{.3840} & \textcolor{blue}{.3894} & .4449 & \textcolor{blue}{.5309} \\
\dsep
\dwiki{} & $P(\beta{=}0.5)$  & .4040 & .4116 & .4368 & .5123 & \textcolor{blue}{.5728} \\
         & \sysname{} & \textcolor{blue}{.4111} & \textcolor{blue}{.4215} & \textcolor{blue}{.4407} & \textcolor{blue}{.5171} & .5727 \\
\bottomrule
\end{tabular}%
}
\caption{Effect of the DPO regularization parameter $\beta$ (NDCG@$K$):
$P(\beta{=}0.5)$ vs.\ \sysname{}.
$P(\beta{=}0.5)$ = \sysname{} with $\beta = 0.5$;
\sysname{} uses the default $\beta = 0.1$.
Blue = better per cell.}
\label{tab:sens-beta}
\end{table}

\begin{table}[t]
\centering
\footnotesize
\resizebox{\columnwidth}{!}{%
\begin{tabular}{llccccc}
\toprule
Dataset & Method & @5 & @10 & @20 & @50 & @100 \\
\midrule
\daw{} & $P(\beta{=}0.5)$  & \textcolor{blue}{.68} & .77 & \textcolor{blue}{.86} & \textcolor{blue}{.86} & \textcolor{blue}{.86} \\
       & \sysname{} & .66 & .77 & .83 & .83 & .83 \\
\dsep
\darctic{} & $P(\beta{=}0.5)$  & \textcolor{blue}{.30} & \textcolor{blue}{.42} & .56 & \textcolor{blue}{.78} & .82 \\
           & \sysname{} & .29 & .41 & \textcolor{blue}{.58} & .77 & .82 \\
\dsep
\dlter{} & $P(\beta{=}0.5)$  & .13 & .20 & .29 & .37 & .40 \\
         & \sysname{} & \textcolor{blue}{.18} & \textcolor{blue}{.25} & \textcolor{blue}{.32} & \textcolor{blue}{.39} & \textcolor{blue}{.43} \\
\dsep
\decir{} & $P(\beta{=}0.5)$  & .01 & .02 & .04 & .08 & \textcolor{blue}{.13} \\
         & \sysname{} & .01 & \textcolor{blue}{.03} & .04 & .08 & \textcolor{blue}{.13} \\
\dsep
\dwtr{} & $P(\beta{=}0.5)$  & .08 & \textcolor{blue}{.15} & .25 & .47 & .65 \\
        & \sysname{} & .08 & \textcolor{blue}{.15} & \textcolor{blue}{.26} & \textcolor{blue}{.48} & \textcolor{blue}{.66} \\
\dsep
\dwiki{} & $P(\beta{=}0.5)$  & .20 & .30 & .42 & .67 & \textcolor{blue}{.83} \\
         & \sysname{} & \textcolor{blue}{.21} & \textcolor{blue}{.32} & .42 & .67 & .81 \\
\bottomrule
\end{tabular}%
}
\caption{Effect of the DPO regularization parameter $\beta$ (Recall@$K$):
$P(\beta{=}0.5)$ vs.\ \sysname{}.
$P(\beta{=}0.5)$ = \sysname{} with $\beta = 0.5$;
\sysname{} uses the default $\beta = 0.1$.
Blue = better per cell.}
\label{tab:sens-beta-recall}
\end{table}

\section{Sensitivity Analysis with Different Numbers of Descriptions}
\label{sec:app-sens-ndesc}

Table~\ref{tab:sens-ndesc} shows NDCG@K and
Table~\ref{tab:sens-ndesc-recall} shows Recall@K for the effect of
increasing the number of candidate descriptions per gold table from 3 to 5.

\begin{table}[t]
\centering
\footnotesize
\resizebox{\columnwidth}{!}{%
\begin{tabular}{llccccc}
\toprule
Dataset & Method & @5 & @10 & @20 & @50 & @100 \\
\midrule
\daw{} & $P(5\text{-}\mathrm{desc})$  & \textcolor{blue}{.8019} & \textcolor{blue}{.8313} & \textcolor{blue}{.8601} & \textcolor{blue}{.8601} & \textcolor{blue}{.8601} \\
       & \sysname{} & .7312 & .7582 & .7869 & .7869 & .7869 \\
\dsep
\darctic{} & $P(5\text{-}\mathrm{desc})$  & .6843 & .6593 & \textcolor{blue}{.7013} & \textcolor{blue}{.7446} & .7645 \\
           & \sysname{} & \textcolor{blue}{.7133} & \textcolor{blue}{.6743} & .6939 & .7416 & \textcolor{blue}{.7678} \\
\dsep
\dlter{} & $P(5\text{-}\mathrm{desc})$  & .6364 & .5987 & .5567 & .4805 & .4694 \\
         & \sysname{} & \textcolor{blue}{.7218} & \textcolor{blue}{.6546} & \textcolor{blue}{.6013} & \textcolor{blue}{.5239} & \textcolor{blue}{.5145} \\
\dsep
\decir{} & $P(5\text{-}\mathrm{desc})$  & .4728 & .4673 & .4832 & \textcolor{blue}{.4897} & \textcolor{blue}{.4731} \\
         & \sysname{} & \textcolor{blue}{.4946} & \textcolor{blue}{.5050} & \textcolor{blue}{.4981} & .4873 & .4672 \\
\dsep
\dwtr{} & $P(5\text{-}\mathrm{desc})$  & .3723 & \textcolor{blue}{.3842} & \textcolor{blue}{.3982} & \textcolor{blue}{.4623} & \textcolor{blue}{.5391} \\
        & \sysname{} & \textcolor{blue}{.3840} & .3840 & .3894 & .4449 & .5309 \\
\dsep
\dwiki{} & $P(5\text{-}\mathrm{desc})$  & \textcolor{blue}{.4328} & \textcolor{blue}{.4364} & \textcolor{blue}{.4550} & \textcolor{blue}{.5248} & \textcolor{blue}{.5834} \\
         & \sysname{} & .4111 & .4215 & .4407 & .5171 & .5727 \\
\bottomrule
\end{tabular}%
}
\caption{Effect of the number of candidate descriptions (NDCG@$K$):
$P(5\text{-}\mathrm{desc})$ vs.\ \sysname{}.
$P(5\text{-}\mathrm{desc})$ = \sysname{} with 5 candidate descriptions
per gold table; \sysname{} has 3 by default.
Blue = better per cell.}
\label{tab:sens-ndesc}
\end{table}

\begin{table}[t]
\centering
\footnotesize
\resizebox{\columnwidth}{!}{%
\begin{tabular}{llccccc}
\toprule
Dataset & Method & @5 & @10 & @20 & @50 & @100 \\
\midrule
\daw{} & $P(5\text{-}\mathrm{desc})$  & \textcolor{blue}{.73} & \textcolor{blue}{.84} & \textcolor{blue}{.90} & \textcolor{blue}{.90} & \textcolor{blue}{.90} \\
       & \sysname{} & .66 & .77 & .83 & .83 & .83 \\
\dsep
\darctic{} & $P(5\text{-}\mathrm{desc})$  & .29 & .41 & \textcolor{blue}{.61} & \textcolor{blue}{.80} & \textcolor{blue}{.84} \\
           & \sysname{} & .29 & .41 & .58 & .77 & .82 \\
\dsep
\dlter{} & $P(5\text{-}\mathrm{desc})$  & .13 & .20 & .28 & .35 & .39 \\
         & \sysname{} & \textcolor{blue}{.18} & \textcolor{blue}{.25} & \textcolor{blue}{.32} & \textcolor{blue}{.39} & \textcolor{blue}{.43} \\
\dsep
\decir{} & $P(5\text{-}\mathrm{desc})$  & .01 & .02 & .04 & .08 & \textcolor{blue}{.13} \\
         & \sysname{} & .01 & \textcolor{blue}{.03} & .04 & .08 & \textcolor{blue}{.13} \\
\dsep
\dwtr{} & $P(5\text{-}\mathrm{desc})$  & .08 & \textcolor{blue}{.16} & \textcolor{blue}{.27} & \textcolor{blue}{.50} & \textcolor{blue}{.67} \\
        & \sysname{} & .08 & .15 & .26 & .48 & .66 \\
\dsep
\dwiki{} & $P(5\text{-}\mathrm{desc})$  & .21 & .32 & \textcolor{blue}{.43} & .66 & \textcolor{blue}{.83} \\
         & \sysname{} & .21 & .32 & .42 & \textcolor{blue}{.67} & .81 \\
\bottomrule
\end{tabular}%
}
\caption{Effect of the number of candidate descriptions (Recall@$K$):
$P(5\text{-}\mathrm{desc})$ vs.\ \sysname{}.
$P(5\text{-}\mathrm{desc})$ = \sysname{} with 5 candidate descriptions
per gold table; \sysname{} has 3 by default.
Blue = better per cell.}
\label{tab:sens-ndesc-recall}
\end{table}

\section{Sensitivity Analysis with Different Numbers of Training Epochs}
\label{sec:app-sens-epochs}

Table~\ref{tab:sens-epochs} shows NDCG@K and
Table~\ref{tab:sens-epochs-recall} shows Recall@K for the effect of
increasing DPO training from 1 epoch to 3 epochs.

\begin{table}[t]
\centering
\footnotesize
\resizebox{\columnwidth}{!}{%
\begin{tabular}{llccccc}
\toprule
Dataset & Method & @5 & @10 & @20 & @50 & @100 \\
\midrule
\daw{} & $P(3\text{-}\mathrm{epoch})$  & .6908 & .7522 & \textcolor{blue}{.7881} & \textcolor{blue}{.7968} & \textcolor{blue}{.7968} \\
       & \sysname{} & \textcolor{blue}{.7312} & \textcolor{blue}{.7582} & .7869 & .7869 & .7869 \\
\dsep
\darctic{} & $P(3\text{-}\mathrm{epoch})$  & \textcolor{blue}{.8080} & \textcolor{blue}{.7131} & \textcolor{blue}{.7059} & \textcolor{blue}{.7595} & \textcolor{blue}{.7951} \\
           & \sysname{} & .7133 & .6743 & .6939 & .7416 & .7678 \\
\dsep
\dlter{} & $P(3\text{-}\mathrm{epoch})$  & .7064 & .6535 & \textcolor{blue}{.6020} & .5105 & .4940 \\
         & \sysname{} & \textcolor{blue}{.7218} & \textcolor{blue}{.6546} & .6013 & \textcolor{blue}{.5239} & \textcolor{blue}{.5145} \\
\dsep
\decir{} & $P(3\text{-}\mathrm{epoch})$  & .4532 & .4813 & .4763 & .4702 & .4584 \\
         & \sysname{} & \textcolor{blue}{.4946} & \textcolor{blue}{.5050} & \textcolor{blue}{.4981} & \textcolor{blue}{.4873} & \textcolor{blue}{.4672} \\
\dsep
\dwtr{} & $P(3\text{-}\mathrm{epoch})$  & .3810 & .3505 & .3571 & .4095 & .4859 \\
        & \sysname{} & \textcolor{blue}{.3840} & \textcolor{blue}{.3840} & \textcolor{blue}{.3894} & \textcolor{blue}{.4449} & \textcolor{blue}{.5309} \\
\dsep
\dwiki{} & $P(3\text{-}\mathrm{epoch})$  & .4069 & .4002 & .4359 & .5142 & \textcolor{blue}{.5753} \\
         & \sysname{} & \textcolor{blue}{.4111} & \textcolor{blue}{.4215} & \textcolor{blue}{.4407} & \textcolor{blue}{.5171} & .5727 \\
\bottomrule
\end{tabular}%
}
\caption{Effect of the number of training epochs (NDCG@$K$):
$P(3\text{-}\mathrm{epoch})$ vs.\ \sysname{}.
$P(3\text{-}\mathrm{epoch})$ = \sysname{} trained for 3 epochs;
\sysname{} uses 1 epoch by default.
Blue = better per cell.}
\label{tab:sens-epochs}
\end{table}

\begin{table}[t]
\centering
\footnotesize
\resizebox{\columnwidth}{!}{%
\begin{tabular}{llccccc}
\toprule
Dataset & Method & @5 & @10 & @20 & @50 & @100 \\
\midrule
\daw{} & $P(3\text{-}\mathrm{epoch})$  & .65 & \textcolor{blue}{.83} & \textcolor{blue}{.91} & \textcolor{blue}{.93} & \textcolor{blue}{.93} \\
       & \sysname{} & \textcolor{blue}{.66} & .77 & .83 & .83 & .83 \\
\dsep
\darctic{} & $P(3\text{-}\mathrm{epoch})$  & \textcolor{blue}{.32} & \textcolor{blue}{.42} & .59 & \textcolor{blue}{.80} & \textcolor{blue}{.86} \\
           & \sysname{} & .29 & .41 & .58 & .77 & .82 \\
\dsep
\dlter{} & $P(3\text{-}\mathrm{epoch})$  & .16 & .23 & .31 & .38 & .42 \\
         & \sysname{} & \textcolor{blue}{.18} & \textcolor{blue}{.25} & \textcolor{blue}{.32} & \textcolor{blue}{.39} & \textcolor{blue}{.43} \\
\dsep
\decir{} & $P(3\text{-}\mathrm{epoch})$  & .01 & .02 & .04 & .08 & .13 \\
         & \sysname{} & .01 & \textcolor{blue}{.03} & .04 & .08 & \textcolor{blue}{.13} \\
\dsep
\dwtr{} & $P(3\text{-}\mathrm{epoch})$  & .08 & .13 & .22 & .42 & .59 \\
        & \sysname{} & .08 & \textcolor{blue}{.15} & \textcolor{blue}{.26} & \textcolor{blue}{.48} & \textcolor{blue}{.66} \\
\dsep
\dwiki{} & $P(3\text{-}\mathrm{epoch})$  & .20 & .28 & \textcolor{blue}{.44} & .67 & \textcolor{blue}{.84} \\
         & \sysname{} & \textcolor{blue}{.21} & \textcolor{blue}{.32} & .42 & .67 & .81 \\
\bottomrule
\end{tabular}%
}
\caption{Effect of the number of training epochs (Recall@$K$):
$P(3\text{-}\mathrm{epoch})$ vs.\ \sysname{}.
$P(3\text{-}\mathrm{epoch})$ = \sysname{} trained for 3 epochs;
\sysname{} uses 1 epoch by default.
Blue = better per cell.}
\label{tab:sens-epochs-recall}
\end{table}

\section{Multi-Field Keyword Search}

\label{sec:kws}

Table~\ref{tab:kws-ndcg} shows NDCG@K and Table~\ref{tab:kws-recall}
shows Recall@K for multi-field \sysname{}, compared to single-field
\sysname{} and AutoDDG.

\begin{table}[t]
\centering
\footnotesize
\resizebox{\columnwidth}{!}{%
\begin{tabular}{llccccc}
\toprule
Dataset & Method & @5 & @10 & @20 & @50 & @100 \\
\midrule
\daw{} & AutoDDG & .5551 & .6192 & .6692 & .6969 & .7030 \\
       & \sysname{} & .7312 & .7582 & .7869 & .7869 & .7869 \\
       & $P(\mathrm{multi})$ & \textcolor{blue}{.8133} & \textcolor{blue}{.8282} & \textcolor{blue}{.8677} & \textcolor{blue}{.8677} & \textcolor{blue}{.8677} \\
\dsep
\darctic{} & AutoDDG & .6274 & .5942 & .5942 & .6360 & .6812 \\
           & \sysname{} & .7133 & .6743 & .6939 & .7416 & .7678 \\
           & $P(\mathrm{multi})$ & \textcolor{blue}{.7303} & \textcolor{blue}{.7146} & \textcolor{blue}{.7330} & \textcolor{blue}{.7755} & \textcolor{blue}{.8058} \\
\dsep
\dlter{} & AutoDDG & .4372 & .4194 & .4011 & .3656 & .3559 \\
         & \sysname{} & .7218 & .6546 & .6013 & .5239 & .5145 \\
         & $P(\mathrm{multi})$ & \textcolor{blue}{.7962} & \textcolor{blue}{.7074} & \textcolor{blue}{.6567} & \textcolor{blue}{.6278} & \textcolor{blue}{.6384} \\
\dsep
\decir{} & AutoDDG & .4215 & .4429 & .4512 & .4408 & .4180 \\
         & \sysname{} & .4946 & .5050 & .4981 & .4873 & .4672 \\
         & $P(\mathrm{multi})$ & \textcolor{blue}{.5575} & \textcolor{blue}{.5347} & \textcolor{blue}{.5049} & \textcolor{blue}{.4965} & \textcolor{blue}{.4757} \\
\dsep
\dwtr{} & AutoDDG & .3748 & .3548 & .3767 & .4159 & .4884 \\
        & \sysname{} & \textcolor{blue}{.3840} & .3840 & .3894 & .4449 & .5309 \\
        & $P(\mathrm{multi})$ & .3802 & \textcolor{blue}{.3786} & \textcolor{blue}{.3858} & \textcolor{blue}{.4657} & \textcolor{blue}{.5912} \\
\dsep
\dwiki{} & AutoDDG & .4139 & .4085 & .4398 & \textcolor{blue}{.5176} & .5811 \\
         & \sysname{} & .4111 & \textcolor{blue}{.4215} & .4407 & .5171 & .5727 \\
         & $P(\mathrm{multi})$ & \textcolor{blue}{.4268} & .4155 & \textcolor{blue}{.4419} & .5172 & \textcolor{blue}{.5849} \\
\bottomrule
\end{tabular}%
}
\caption{Multi-field KWS (NDCG@$K$):
AutoDDG vs.\ \sysname{} vs.\ $P(\mathrm{multi})$.
$P(\mathrm{multi})$ = \sysname{} using multiple fields,
\sysname{} = single field (description),
AutoDDG = single field (description).
Blue = best per cell.}
\label{tab:kws-ndcg}
\end{table}

\begin{table}[t]
\centering
\footnotesize
\resizebox{\columnwidth}{!}{%
\begin{tabular}{llccccc}
\toprule
Dataset & Method & @5 & @10 & @20 & @50 & @100 \\
\midrule
\daw{} & AutoDDG & .49 & .64 & .76 & .83 & .85 \\
       & \sysname{} & .66 & .77 & .83 & .83 & .83 \\
       & $P(\mathrm{multi})$ & \textcolor{blue}{.75} & \textcolor{blue}{.83} & \textcolor{blue}{.92} & \textcolor{blue}{.92} & \textcolor{blue}{.92} \\
\dsep
\darctic{} & AutoDDG & .24 & .34 & .48 & .68 & .77 \\
           & \sysname{} & .29 & .41 & .58 & .77 & .82 \\
           & $P(\mathrm{multi})$ & \textcolor{blue}{.32} & \textcolor{blue}{.46} & \textcolor{blue}{.63} & \textcolor{blue}{.82} & \textcolor{blue}{.88} \\
\dsep
\dlter{} & AutoDDG & .08 & .14 & .22 & .32 & .35 \\
         & \sysname{} & .18 & .25 & .32 & .39 & .43 \\
         & $P(\mathrm{multi})$ & \textcolor{blue}{.24} & \textcolor{blue}{.31} & \textcolor{blue}{.39} & \textcolor{blue}{.53} & \textcolor{blue}{.61} \\
\dsep
\decir{} & AutoDDG & .01 & .02 & .04 & .08 & .12 \\
         & \sysname{} & .01 & \textcolor{blue}{.03} & .04 & .08 & \textcolor{blue}{.13} \\
         & $P(\mathrm{multi})$ & \textcolor{blue}{.01} & .02 & \textcolor{blue}{.04} & \textcolor{blue}{.08} & \textcolor{blue}{.13} \\
\dsep
\dwtr{} & AutoDDG & .08 & .14 & .25 & .43 & .59 \\
        & \sysname{} & .08 & .15 & .26 & .48 & .66 \\
        & $P(\mathrm{multi})$ & \textcolor{blue}{.09} & \textcolor{blue}{.16} & \textcolor{blue}{.27} & \textcolor{blue}{.52} & \textcolor{blue}{.78} \\
\dsep
\dwiki{} & AutoDDG & \textcolor{blue}{.21} & .29 & \textcolor{blue}{.41} & \textcolor{blue}{.66} & .83 \\
         & \sysname{} & \textcolor{blue}{.21} & \textcolor{blue}{.32} & \textcolor{blue}{.42} & .67 & .81 \\
         & $P(\mathrm{multi})$ & .20 & .29 & .41 & .64 & \textcolor{blue}{.84} \\
\bottomrule
\end{tabular}%
}
\caption{Multi-field KWS (Recall@$K$):
AutoDDG vs.\ \sysname{} vs.\ $P(\mathrm{multi})$.
$P(\mathrm{multi})$ = \sysname{} using multiple fields,
\sysname{} = single field (description),
AutoDDG = single field (description).
Blue = best per cell.}
\label{tab:kws-recall}
\end{table}

Table~\ref{tab:kws-ablation} shows NDCG@K for the effect of removing
each of the three fields (description, table/column name expansions,
and table context) from multi-field \sysname{}.

\begin{table}[t]
\centering
\footnotesize
\resizebox{\columnwidth}{!}{%
\begin{tabular}{llccccc}
\toprule
Dataset & Method & @5 & @10 & @20 & @50 & @100 \\
\midrule
\daw{} & $P(\mathrm{multi})$ & \textcolor{blue}{.8133} & \textcolor{blue}{.8282} & \textcolor{blue}{.8677} & \textcolor{blue}{.8677} & \textcolor{blue}{.8677} \\
       & ${-}\mathrm{desc}$ & .7594 & .7668 & .7895 & .7895 & .7895 \\
       & ${-}\mathrm{exp}$ & .7312 & .7582 & .7869 & .7869 & .7869 \\
\dsep
\darctic{} & $P(\mathrm{multi})$ & .7303 & \textcolor{blue}{.7146} & \textcolor{blue}{.7330} & \textcolor{blue}{.7755} & \textcolor{blue}{.8058} \\
           & ${-}\mathrm{desc}$ & \textcolor{blue}{.7602} & .7162 & .6726 & .7139 & .7255 \\
           & ${-}\mathrm{exp}$ & .7133 & .6743 & .6939 & .7416 & .7678 \\
\dsep
\dlter{} & $P(\mathrm{multi})$ & \textcolor{blue}{.7962} & \textcolor{blue}{.7074} & \textcolor{blue}{.6567} & \textcolor{blue}{.6278} & \textcolor{blue}{.6384} \\
         & ${-}\mathrm{desc}$ & .5938 & .5639 & .5236 & .5084 & .4959 \\
         & ${-}\mathrm{exp}$ & .7218 & .6546 & .6013 & .5239 & .5145 \\
\dsep
\decir{} & $P(\mathrm{multi})$ & \textcolor{blue}{.5575} & \textcolor{blue}{.5347} & \textcolor{blue}{.5049} & .4965 & \textcolor{blue}{.4757} \\
         & ${-}\mathrm{desc}$ & .4478 & .4737 & .4474 & .4614 & .4422 \\
         & ${-}\mathrm{exp}$ & .5053 & .5231 & .5159 & \textcolor{blue}{.4996} & .4770 \\
         & ${-}\mathrm{ctx}$ & .5474 & .5333 & .4925 & .4816 & .4548 \\
\dsep
\dwtr{} & $P(\mathrm{multi})$ & .3802 & .3786 & .3858 & .4657 & \textcolor{blue}{.5912} \\
        & ${-}\mathrm{desc}$ & .3698 & .3734 & .3811 & .4632 & .5844 \\
        & ${-}\mathrm{exp}$ & \textcolor{blue}{.4053} & \textcolor{blue}{.4003} & \textcolor{blue}{.4150} & \textcolor{blue}{.4816} & .5993 \\
        & ${-}\mathrm{ctx}$ & .3583 & .3581 & .3559 & .4296 & .5254 \\
\dsep
\dwiki{} & $P(\mathrm{multi})$ & \textcolor{blue}{.4268} & \textcolor{blue}{.4155} & \textcolor{blue}{.4419} & .5172 & \textcolor{blue}{.5849} \\
         & ${-}\mathrm{desc}$ & .3179 & .3187 & .3476 & .4207 & .4946 \\
         & ${-}\mathrm{exp}$ & .4214 & .4146 & .4403 & \textcolor{blue}{.5233} & .5812 \\
         & ${-}\mathrm{ctx}$ & .4168 & .4135 & .4369 & .5091 & .5706 \\
\bottomrule
\end{tabular}%
}
\caption{Field ablation (NDCG@$K$) for multi-field $P(\mathrm{multi})$.
${-}\mathrm{desc}$ = w/o description,
${-}\mathrm{exp}$ = w/o table/column name expansions,
${-}\mathrm{ctx}$ = w/o table context (\decir{}, \dwtr{}, \dwiki{} only).
Blue = best per cell.}
\label{tab:kws-ablation}
\end{table}